\documentclass{article}

\PassOptionsToPackage{numbers, compress}{natbib}

\usepackage[preprint]{neurips_2026}

\usepackage{authblk}

\usepackage{microtype}
\usepackage{graphicx}
\usepackage{subcaption}
\usepackage{booktabs} 
\usepackage{multirow}
\usepackage[table]{xcolor}
\usepackage{tabularx}
\usepackage{enumitem}
\usepackage{caption}
\usepackage{float}
\usepackage{hyperref}

\usepackage{algorithm}
\usepackage{algorithmic}

\usepackage{pgfplots}
\pgfplotsset{compat=1.18}

\usepackage{amsmath}
\usepackage{amssymb}
\usepackage{mathtools}
\usepackage{amsthm}

\usepackage{tikz}
\usepackage{arydshln}

\definecolor{myred}{RGB}{200, 0, 0}
\definecolor{myblue}{RGB}{0, 0, 200}
\definecolor{mypurple}{RGB}{128, 0, 128}

\newcolumntype{Y}{>{\centering\arraybackslash}X}

\usepackage[capitalize,noabbrev]{cleveref}

\theoremstyle{plain}

\theoremstyle{definition}

\theoremstyle{remark}

\usepackage[textsize=tiny]{todonotes}

\title{SIMON: Saliency-aware Integrative Multi-view Object-centric Neural Decoding}

\author[1]{YuSheng Lin}
\author[1]{Ji-Hwa Tsai}
\author[1]{Chun-Shu Wei}
\affil[1]{National Yang Ming Chiao Tung University, Taiwan}

\date{}

\begin{document}

\maketitle

\begin{abstract}
Recent EEG-to-image retrieval methods leverage pretrained vision encoders and foveation-inspired priors, but typically assume a fixed, center-focused view. This center bias conflicts with content-driven human attention, creating a geometric–semantic dissociation between visual features and EEG responses. We propose \textbf{SIMON}, a saliency-aware multi-view framework for zero-shot EEG-to-image retrieval. SIMON combines foreground segmentation and saliency prediction to select fixation centers via Saliency-Aware Sampling (SAS), then generates foveated views that emphasize informative object regions while suppressing background clutter. On THINGS-EEG, SIMON achieves state-of-the-art performance in both intra- and inter-subject settings, reaching an average Top-1 accuracy of 69.7\% and 19.6\% respectively, consistently outperforming recent competitive baselines. Analyses across sampling granularity, EEG channel topology, and visual/brain encoder backbones further support the robustness of saliency-aware multi-view integration. Our code and models are publicly available at \url{https://github.com/simonlink666/SIMON}.
\end{abstract}



\section{Introduction}
\label{sec:intro}

Decoding visual information from non-invasive recordings of human brain activity is an important problem in both computational neuroscience and brain--computer interface (BCI) research. With the rise of deep generative models and large-scale pretraining \citep{radford2021learning}, EEG-based visual decoding has progressed from recovering low-level visual cues toward retrieving high-level semantic content \citep{song2024decoding, li2024visual, chen2024visual}.

Despite these advances, an important mismatch remains in how visual stimuli are represented for EEG decoding. Recent methods often incorporate foveation-inspired priors, such as center-focused blur, to mimic the non-uniform spatial resolution of the human visual system \citep{wu2025bridging}. However, they still typically assume that the most informative region lies near the geometric center of the image. This assumption is often inaccurate for natural images, where semantically important objects may appear away from the center. As a result, models may preserve irrelevant central background content while attenuating informative foreground regions in the periphery, weakening the alignment between visual features and EEG responses.

\begin{figure}[t!] 
    \centering
    \begin{minipage}[t]{0.48\textwidth}
        \centering
        \includegraphics[width=0.8\linewidth]{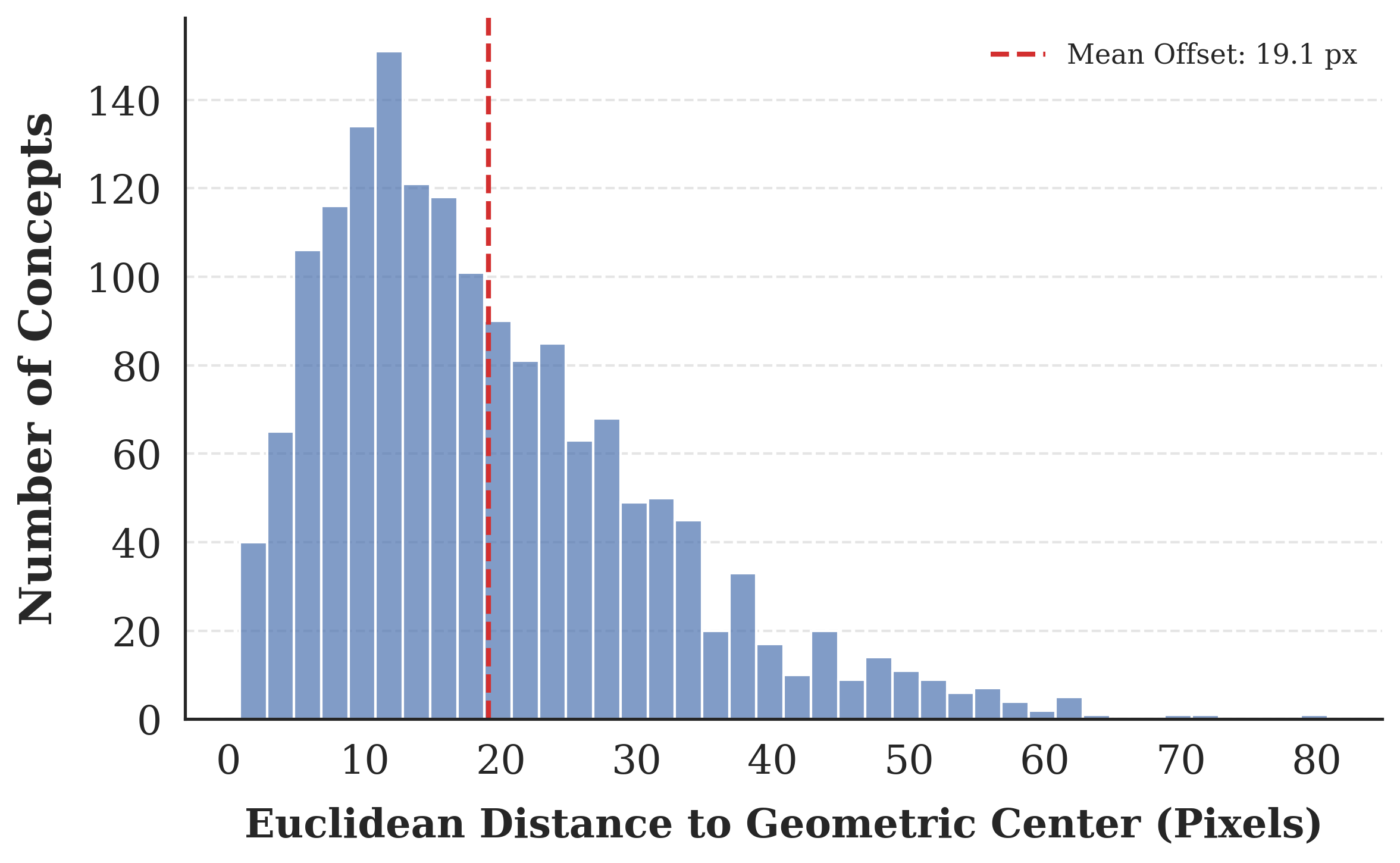}
        
        \vspace{2pt}
        \small (a)
    \end{minipage}%
    \hfill 
    \begin{minipage}[t]{0.48\textwidth}
        \centering
        \includegraphics[width=0.8\linewidth]{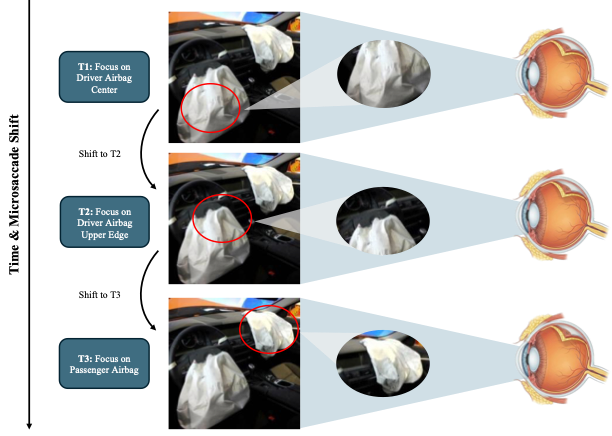}
        
        \vspace{2pt}
        \small (b)
    \end{minipage}
    
    \caption{(a) Quantitative Assessment of Geometric-Semantic Dissociation; (b) Illustration of the Saliency-Aware Sampling.}
    \label{fig:combined_initial_analysis}

    \vspace{-0.5cm} 
\end{figure}

We refer to this mismatch as \textbf{Geometric--Semantic Dissociation}: the discrepancy between the \emph{geometric focus} imposed by a center-fixed visual prior and the \emph{semantic focus} that more closely reflects content-driven human attention. Human vision is inherently active and content-driven \citep{rensink2000dynamic}; our eyes naturally shift toward semantically salient objects rather than remaining fixed at geometric coordinates \citep{noton1971scanpaths}. To quantify this effect, we analyze the spatial distribution of semantic focus on the THINGS dataset. As shown in \Cref{fig:combined_initial_analysis}(a), the semantic centroid, computed from saliency-weighted structure, exhibits a mean displacement of 19.1 pixels from the image center across 1,654 concept images. This systematic offset indicates that rigid center-focused encoding often fails to capture the regions most relevant to the corresponding EEG activity.

Motivated by this observation, we propose \textbf{SIMON}, a \textbf{S}aliency-aware \textbf{I}ntegrative \textbf{M}ulti-view \textbf{O}bject-centric \textbf{N}eural decoding framework for zero-shot EEG-to-image retrieval. Instead of using a single center-fixed view, SIMON first estimates foreground structure and visual saliency using BiRefNet \citep{zheng2024birefnet} and Saliency Unification through Mamba (SUM) \citep{hosseini2025sum}. It then applies \textbf{Saliency-Aware Sampling (SAS)} to select fixation centers that cover semantically informative regions, followed by foveated view generation and multi-view aggregation. This design is consistent with the active nature of human vision: even under brief RSVP presentations, covert attention rapidly prioritizes salient content \citep{posner1980orienting, carrasco2011visual}, while microsaccadic behavior supports localized sampling of fine details \citep{hafed2002microsaccades, martinez2004role}.

We evaluate SIMON on the THINGS-EEG benchmark \citep{gifford2022large}. Beyond achieving state-of-the-art performance in both intra- and inter-subject retrieval, we show that SIMON is particularly effective when semantic and geometric centers diverge. We further analyze the effects of sampling granularity, EEG channel topology, and different visual and brain encoder backbones, demonstrating the robustness of saliency-aware multi-view integration.

Our contributions are three-fold:
\begin{itemize}[leftmargin=*, nosep]
    \item \textbf{Characterizing geometric-semantic dissociation.} We identify and quantify a practically important mismatch in center-biased EEG-to-image retrieval, where the geometric center does not necessarily match the semantically informative region.
    \item \textbf{A saliency-aware multi-view framework.} We propose \textbf{SIMON}, which combines dual-source saliency extraction, saliency-aware sampling, and multi-view foveated encoding to build object-centric visual representations better aligned with EEG responses.
    \item \textbf{Comprehensive evaluation.} We show that SIMON achieves state-of-the-art EEG-to-image retrieval performance and analyze its behavior across dissociation severity, sampling granularity, EEG topology, and encoder backbones.
\end{itemize}

\begin{figure}[b]
    \centering
    \vspace{-0.4cm}
    \includegraphics[width=0.7\linewidth]{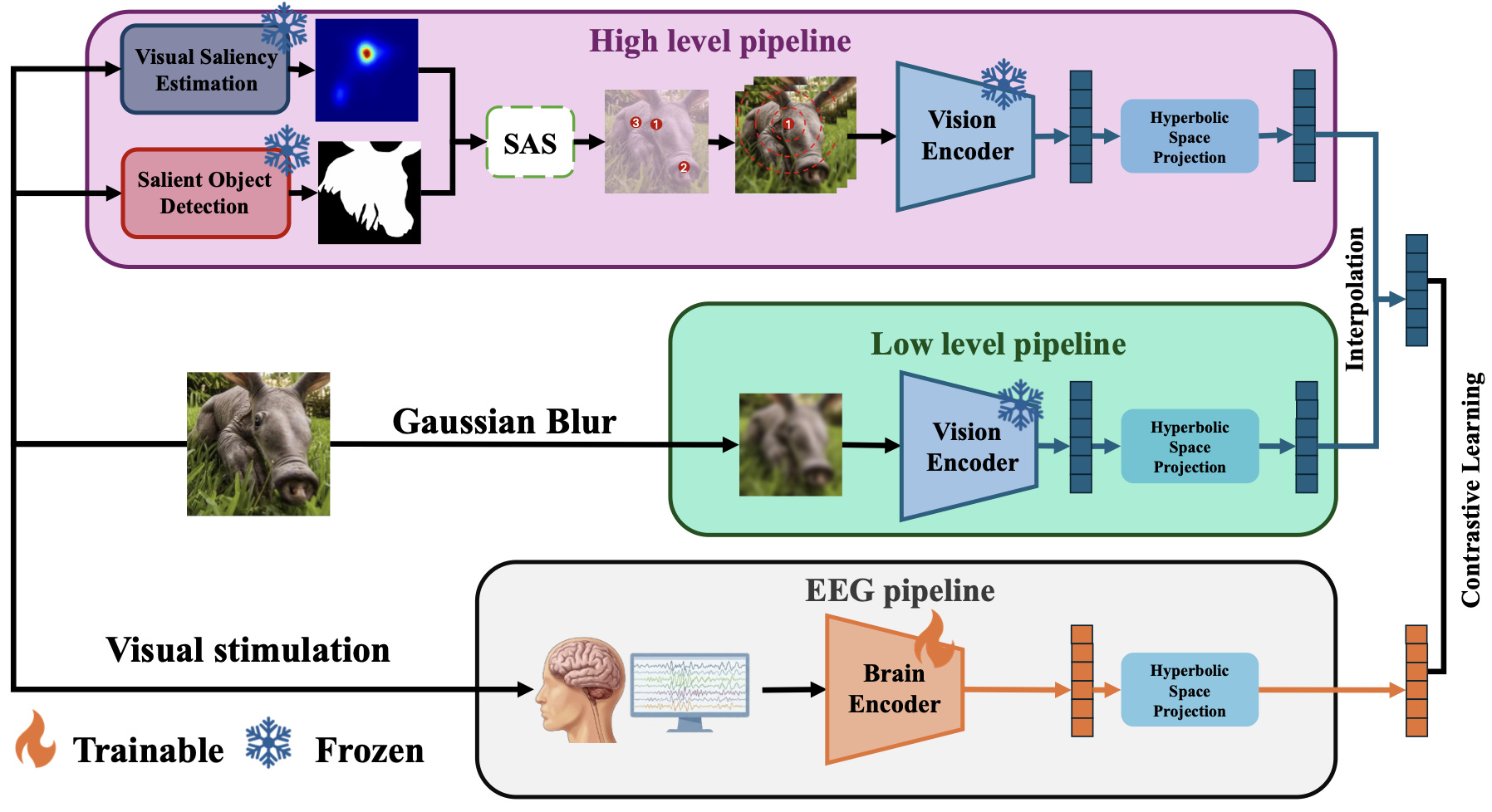}
    \caption{\textbf{An overview of the proposed SIMON framework.} The pipeline integrates visual saliency estimation with a Saliency-Aware Sampling (SAS) strategy. The selected high-resolution crops are processed by a vision encoder and projected into hyperbolic space to align with EEG embeddings via contrastive learning.}
    \label{fig:framework}
    \vspace{-0.5cm} 
\end{figure}


\section{Related Work}

\subsection{Semantic and Spatial Mechanisms in Visual Processing}
Visual neuroscience has long emphasized the distinction between \emph{what} is represented and \emph{where} it is represented, as well as the role of object-based and content-driven attention in perception \citep{cavanagh2023architecture, klindt2017neural, rolls2024two}. These studies suggest that visual processing depends on the interaction between semantic content and spatially localized selection. However, this line of work has mainly focused on single-neuron encoding models or high-spatial-resolution fMRI settings, rather than EEG-based visual decoding, where spatial precision is limited and visual representations must be inferred from coarse neural measurements.

\subsection{Neural Decoding of Saliency and Visual Priority}
A related line of work attempts to decode saliency maps, gaze patterns, or spatial priority directly from neural activity \citep{khaleghi2022visual, oconnell2018predicting, palazzo2021decoding}. These studies show that macroscopic brain signals contain information about attentional allocation and spatial emphasis. In most cases, however, saliency is treated as the decoding target itself, rather than as a front-end mechanism for constructing better visual features before cross-modal alignment. Our work differs in that saliency is used to guide visual representation learning for EEG-to-image retrieval.

\subsection{Cross-Modal Alignment and Foveated Priors in EEG Decoding}
Recent EEG-based visual decoding methods increasingly rely on cross-modal alignment between brain signals and pretrained visual representations. Early approaches such as \textbf{NICE} \citep{song2024decoding} and \textbf{ATM} \citep{li2024visual} align EEG with image features through contrastive or transformer-based frameworks, while later methods such as \textbf{VE-SDN} \citep{chen2024visual} further refine semantic correspondence. Other works enrich visual priors through additional cues, such as color-aware segmentation in \textbf{CUBE} \citep{akbarinia2025footprint} or perceptual variability modeling in \textbf{NeuroBridge} \citep{zhang2025neurobridge}. 

A related direction introduces foveation-inspired priors into EEG decoding. For example, \citet{wu2025bridging} use uncertainty-aware foveated blur to model peripheral resolution decay. While such priors are biologically motivated, they still typically rely on stochastic or center-fixed view selection. Our work addresses this limitation by replacing center-fixed visual encoding with saliency-aware multi-view sampling and object-centric representation construction.

\subsection{Hyperbolic Representation Learning}
Because neural signals and semantic concepts often exhibit hierarchical structure, several recent multimodal methods have adopted hyperbolic geometry for representation learning. \textbf{HyFI} \citep{jo2026hyfi} shows that hyperbolic contrastive learning can improve multimodal retrieval by providing a more structured latent space. Following this line of work, we use hyperbolic alignment as the shared space for EEG and visual embeddings, while focusing our main methodological contribution on the visual-side design that improves semantic alignment before cross-modal matching.

\section{Method}
\label{sec:method}

\subsection{Overall Framework}
We address the task of zero-shot brain-to-image retrieval, aiming to align EEG signals evoked by visual stimuli with their corresponding semantic image representations. Our framework, named \textbf{SIMON}, adopts a dual-stream architecture that projects both brain and visual modalities into a shared hyperbolic space.

While prior methods often utilize whole-image embeddings that suffer from background clutter, we propose a \textbf{Saliency-Aware Multi-View Visual Encoding} strategy. Our approach leverages a dual-source mechanism combining salient object detection (BiRefNet) and saliency prediction (SUM) to generate diverse, noise-suppressed foveated views. These views are aggregated and aligned with EEG features via geodesic interpolation in the Lorentz model.

\subsection{Saliency-Aware Multi-View Visual Encoding}
To construct a visual representation that is both semantically robust and aligned with neural attention, we implement a three-stage encoding process: Dual-Source Saliency Extraction, Saliency-Aware Sampling, and Foveated View Generation.

\textbf{Dual-Source Saliency Extraction.} Unlike standard encoders that treat all pixels equally, we argue that EEG signals are primarily correlated with the semantic subject rather than background noise. To accurately dissociate the subject from the background while identifying high-attention regions, we employ two specialized pre-trained models:
\begin{enumerate}[leftmargin=*, nosep]
    \item \textbf{Salient Object Detection:} We utilize BiRefNet to generate a precise foreground alpha matte, denoted as $M_{fg} \in [0, 1]^{H \times W}$. BiRefNet excels at capturing high-fidelity boundary details, ensuring that the semantic object is cleanly separated from the background.
    \item \textbf{Visual Saliency Estimation:} Simultaneously, we employ the SUM network to estimate a continuous saliency map $S_{att} \in [0, 1]^{H \times W}$. This map quantifies the pixel-wise probability of attracting human visual attention.
\end{enumerate}
These two maps are combined to guide our sampling strategy, ensuring that fixation points are located within the semantic object ($M_{fg}$) but prioritized by their visual salience ($S_{att}$).

\begin{figure}[htbp] 
    \centering
    \includegraphics[width=0.7\linewidth]{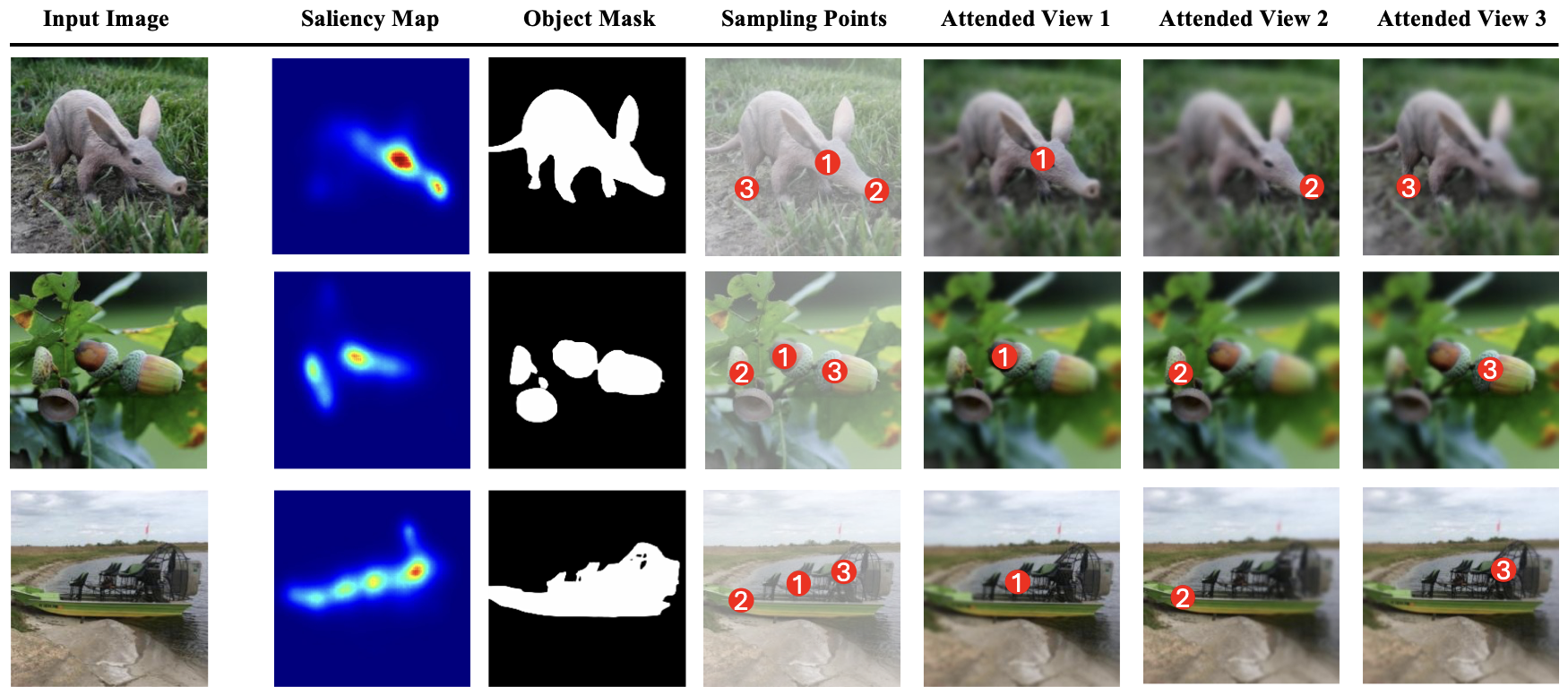}
    \caption{\textbf{Visualization of the saliency-aware multi-view generation}
    From left to right: input image, semantic saliency map, foreground mask, sampled fixation centers, and the resulting foveated views. The sampled centers adapt to off-center semantic regions, allowing SIMON to preserve informative foreground details, such as the extremities of the aardvark and the elongated structure of the airboat, that are often suppressed by center-fixed foveation.}
    \label{fig:view_generation_pipeline}
\end{figure}

\textbf{Saliency-Aware Sampling.} To simulate the human visual exploration process (saccades), we propose a Saliency-Aware Sampling (SAS) strategy. Instead of random sampling, we iteratively select $K$ fixation centers $\mathcal{C} = \{c_1, \dots, c_K\}$ based on the interaction between $M_{fg}$ and $S_{att}$.

First, we determine an initial seed point $c_1$ by calculating the centroid of the saliency map $S_{att}$ masked by the foreground $M_{fg}$:
\begin{equation}
    c_1 = \text{Centroid}(S_{att} \odot \mathbb{I}(M_{fg} > \tau))
\end{equation}
where $\mathbb{I}(\cdot)$ is the indicator function and $\tau$ is a threshold.

Subsequent points are selected via SAS. For a candidate pixel $p$ within the foreground, the selection score is proportional to its saliency intensity $S_{att}(p)$ weighted by its distance to existing centers. This ensures that the sampled views cover diverse, high-saliency parts of the object (e.g., distinguishing the head and torso) while ignoring the background. The detailed procedure is outlined in Algorithm~\ref{alg:sampling}.

\begin{figure}[tb] 
    \centering
    \begin{minipage}{0.7\linewidth} 
        \vspace{-1.0cm} 
        \begin{algorithm}[H] 
           \caption{Saliency-Aware Sampling}
           \label{alg:sampling}
        \begin{algorithmic}
           \STATE {\bfseries Input:} Saliency map $S_{att}$, Foreground mask $M_{fg}$, Views $K$
           \STATE {\bfseries Output:} Fixation centers $\mathcal{C}$
           \STATE Mask candidate region $\Omega = \{p \mid M_{fg}(p) > \tau\}$
           \STATE Find seed $c_1 \leftarrow \text{Centroid of } S_{att} \text{ within } \Omega$
           \STATE $\mathcal{C} \leftarrow \{c_1\}$
           \FOR{$k=2$ {\bfseries to} $K$}
           \STATE Compute distance $D(p) = \min_{c_j \in \mathcal{C}} \|p - c_j\|_2, \forall p \in \Omega$
           \STATE Compute score $J(p) = D(p) \cdot (S_{att}(p))^\gamma$
           \STATE Select $c_k \leftarrow \arg\max_{p \in \Omega} J(p)$
           \STATE $\mathcal{C} \leftarrow \mathcal{C} \cup \{c_k\}$
           \ENDFOR
        \end{algorithmic}
        \end{algorithm}
    \end{minipage}
\end{figure}

\textbf{Foveated View Generation.} For each selected center $c_k$, we generate a foveated view $v_k$ to simulate the human visual focus. Our generation process consists of two steps: background suppression and radial foveation.

First, to mitigate background noise while preserving the semantic integrity of the object, we construct a base image $I_{base}$. We apply a global background blur $I_{bg\_blur}$ and composite the sharp original image $I_{orig}$ using the BiRefNet mask $M_{fg}$:
\begin{equation}
    I_{base} = I_{orig} \odot M_{fg} + I_{bg\_blur} \odot (1 - M_{fg})
\end{equation}
This ensures that the background clutter is suppressed before view generation.

Next, for each fixation center $c_k$, we apply a distance-dependent radial Gaussian blur to $I_{base}$. The blur intensity increases with the distance from $c_k$, simulating the resolution decay of the human retina from the fovea to the periphery. The final view $v_k$ is obtained as:
\begin{equation}
    v_k(p) = \text{Blur}(I_{base}, \sigma(\|p - c_k\|_2))
\end{equation}
where $\sigma(\cdot)$ defines the blur radius function. This results in views where the fixated region is sharp, while peripheral regions (even within the object) and the background are progressively blurred.

Finally, the $K$ views are encoded by a pre-trained CLIP encoder $f_v$ and aggregated via mean pooling to form the final semantic embedding $\mathbf{z}_v^s$:
\begin{equation}
    \mathbf{z}_v^s = \frac{1}{K} \sum_{k=1}^{K} \frac{f_v(v_k)}{\|f_v(v_k)\|_2}
\end{equation}

\subsection{EEG Encoding and Hyperbolic Alignment}
We adopt a dual-stream design to align saliency-aware visual embeddings with EEG in a shared hyperbolic space.

\textbf{Hyperbolic Projection.} We use the EEGProject encoder $f_b$ to extract EEG features from $x_b$ \cite{jo2026hyfi} and align them with aggregated visual embeddings $\mathbf{z}_v^s$ in the $n$-dimensional Lorentz model $\mathbb{L}^n$. Both modalities are mapped into $\mathbb{L}^n$ via the exponential map at the origin:
\begin{equation}
\mathbf{h}_v^s=\exp_O^\kappa(\alpha_v W_v \mathbf{z}_v^s),\quad
\mathbf{h}_b=\exp_O^\kappa(\alpha_b f_b(x_b))
\end{equation}
where $W_v$ is learnable and $\alpha_v,\alpha_b$ control embedding norms.

\textbf{Geodesic Interpolation and Objective.} We form a target visual embedding by geodesically interpolating between semantic and perceptual embeddings ($\mathbf{h}_v^s$, $\mathbf{h}_v^p$):
\begin{equation}
\hat{\mathbf{h}}_v=\gamma_{\mathbf{h}_v^s\to \mathbf{h}_v^p}(t),\ \ t\in[0,1].
\end{equation}
Training uses a symmetric hyperbolic InfoNCE loss with similarity $s(\mathbf{u},\mathbf{v})=-d_{\mathbb{L}}^2(\mathbf{u},\mathbf{v})$:
\begin{equation}
\mathcal{L}_{total}=\frac{1}{2B}\sum_{i=1}^{B}\left(\mathcal{L}_{v\to b,i}+\mathcal{L}_{b\to v,i}\right),
\quad
\end{equation}
\begin{equation}
\mathcal{L}_{v\to b,i}=-\log\frac{\exp(\lambda s(\hat{\mathbf{h}}_{v,i},\mathbf{h}_{b,i}))}{\sum_{j=1}^{B}\exp(\lambda s(\hat{\mathbf{h}}_{v,i},\mathbf{h}_{b,j}))}.
\end{equation}

\section{Experiments}
\label{sec:experiments}

\subsection{Experimental Setup}
\label{sec:setup}

We evaluate on THINGS-EEG~\cite{gifford2022large}, which contains paired image--EEG recordings from 10 subjects. Following prior work~\cite{song2024decoding, wu2025bridging}, we extract the $[0,1000]$ ms post-stimulus window, apply baseline correction using the preceding 200 ms, resample EEG signals to 250 Hz, and average trials of the same image.

SIMON is implemented in PyTorch and trained with AdamW~\cite{loshchilov2019decoupled}. For intra-subject retrieval, we use EEGProject~\cite{wu2025bridging} with 17 occipital--parietal channels; for inter-subject retrieval, we use TSconv~\cite{song2024decoding} with 63 channels. Visual features are extracted with OpenCLIP RN50~\cite{cherti2023reproducible, he2016deep}, and BiRefNet~\cite{zheng2024birefnet} together with SUM~\cite{hosseini2025sum} is used to generate $K=3$ saliency-aware foveated views. Full implementation details are provided in Appendix~A.

\subsection{Performance Evaluation}
\label{sec:performance}

Table~\ref{tab:results} compares SIMON with recent state-of-the-art methods on THINGS-EEG. Our method achieves the best overall performance in both intra-subject and inter-subject retrieval.

\begin{table}[ht]
\centering
\caption{Comparison results on the THINGS-EEG dataset. Bold and underlined values indicate the best and second-best results, respectively.}
\label{tab:results}
\resizebox{\textwidth}{!}{%
\begin{tabular}{l *{22}{c}}
\toprule
\multirow{2}{*}{Method}& \multicolumn{2}{c}{Subject 1} & \multicolumn{2}{c}{Subject 2} & \multicolumn{2}{c}{Subject 3} & \multicolumn{2}{c}{Subject 4} & \multicolumn{2}{c}{Subject 5} & \multicolumn{2}{c}{Subject 6} & \multicolumn{2}{c}{Subject 7} & \multicolumn{2}{c}{Subject 8} & \multicolumn{2}{c}{Subject 9} & \multicolumn{2}{c}{Subject 10} & \multicolumn{2}{c}{Average} \\
\cmidrule(lr){2-3} \cmidrule(lr){4-5} \cmidrule(lr){6-7} \cmidrule(lr){8-9} \cmidrule(lr){10-11} \cmidrule(lr){12-13} \cmidrule(lr){14-15} \cmidrule(lr){16-17} \cmidrule(lr){18-19} \cmidrule(lr){20-21} \cmidrule(lr){22-23}
& T-1 & T-5 & T-1 & T-5 & T-1 & T-5 & T-1 & T-5 & T-1 & T-5 & T-1 & T-5 & T-1 & T-5 & T-1 & T-5 & T-1 & T-5 & T-1 & T-5 & T-1 & T-5 \\
\midrule
\midrule
\multicolumn{23}{c}{\textbf{Intra-subject: train and test on one subject}} \\
\midrule
BraVL   & 6.1 & 17.9 & 4.9 & 14.9 & 5.6 & 17.4 & 5.0 & 15.1 & 4.0 & 13.4 & 6.0 & 18.2 & 6.5 & 20.4 & 8.8 & 23.7 & 4.3 & 14.0 & 7.0 & 19.7 & 5.8 & 17.5 \\
NICE    & 13.2 & 39.5 & 13.5 & 40.3 & 14.5 & 42.7 & 20.6 & 52.7 & 10.1 & 31.5 & 16.5 & 44.0 & 17.0 & 42.1 & 22.9 & 56.1 & 15.4 & 41.6 & 17.4 & 45.8 & 16.1 & 43.6 \\
ATM-S   & 25.6 & 60.4 & 22.0 & 54.5 & 25.0 & 62.4 & 31.4 & 60.9 & 12.9 & 43.0 & 21.3 & 51.1 & 30.5 & 61.5 & 38.8 & 72.0 & 30.4 & 51.5 & 29.1 & 63.5 & 28.5 & 60.4 \\
Cog-cap & 31.4 & 79.7 & 31.4 & 77.8 & 38.2 & 85.7 & 40.4 & 85.8 & 24.4 & 66.3 & 34.8 & 78.8 & 34.7 & 81.0 & 48.1 & 88.6 & 37.4 & 79.4 & 35.6 & 79.3 & 35.6 & 80.2 \\
UBP     & 41.2 & 70.5 & 51.2 & 80.9 & 51.2 & 82.0 & 51.1 & 76.9 & 42.2 & 72.8 & 57.5 & 83.5 & 49.0 & 79.9 & 58.6 & 85.8 & 45.1 & 76.2 & 61.5 & 88.2 & 50.9 & 79.7 \\
NeuroBridge & 50.0 & 77.6 & 63.2 & 90.6 & 61.6 & 91.1 & 61.4 & \underline{90.0} & 54.8 & 85.0 & 69.7 & 92.9 & 62.7 & 88.8 & 71.2 & 95.1 & \underline{64.0} & \textbf{91.0} & 73.6 & \textbf{97.1} & 63.2 & 89.9 \\
HyFI & \underline{60.6} & \underline{85.3} & \underline{65.9} & \underline{94.0} & \underline{69.5} & \textbf{93.9} & \underline{66.5} & 89.8 & \underline{55.0} & \underline{86.0} & \underline{74.4} & \textbf{95.0} & \textbf{68.4} & \underline{91.3} & \textbf{78.9} & \underline{96.9} & \textbf{66.0} & \underline{90.6} & \underline{77.0} & 96.4 & \underline{68.2} & \underline{91.9} \\
\textbf{SIMON} & \textbf{62.8} & \textbf{87.2} & \textbf{69.0} & \textbf{95.2} & \textbf{71.0} & \textbf{94.2} & \textbf{69.8} & \textbf{93.3} & \textbf{59.5} & \textbf{88.0} & \textbf{75.5} & \textbf{95.3} & \underline{66.8} & \textbf{93.3} & \underline{78.3} & \textbf{97.7} & 62.5 & 88.5 & \textbf{81.8} & \underline{96.7} & \textbf{69.7} & \textbf{92.9} \\
\midrule
\midrule
\multicolumn{23}{c}{\textbf{Inter-subject: leave one subject out for test}} \\
\midrule
BraVL & 2.3 & 8.0 & 1.5 & 6.3 & 1.9 & 6.7 & 2.1 & 8.1 & 2.2 & 7.6 & 1.6 & 6.4 & 2.3 & 8.5 & 1.8 & 7.0 & 1.4 & 5.9 & 1.7 & 6.7 & 1.5 & 5.6 \\
NICE & 7.6 & 22.8 & 5.9 & 20.5 & 6.0 & 22.3 & 6.3 & 20.7 & 4.4 & 18.3 & 5.6 & 22.2 & 5.6 & 19.7 & 6.3 & 22.0 & 5.7 & 17.6 & 8.4 & 28.3 & 6.2 & 21.4 \\
NICE-G & 5.9 & 21.4 & 6.4 & 22.7 & 5.5 & 20.1 & 6.1 & 21.0 & 4.7 & 19.5 & 6.2 & 22.5 & 5.9 & 19.1 & 7.3 & 25.3 & 6.2 & 18.3 & 6.2 & 26.3 & 5.9 & 21.6 \\
ATM-S & 10.5 & 26.8 & 7.1 & 24.8 & 11.9 & 33.8 & 14.7 & 39.4 & 7.0 & 23.9 & 11.1 & 35.8 & \underline{16.1} & 43.5 & 15.0 & 40.3 & 4.9 & 22.7 & 20.5 & 46.5 & 11.8 & 33.7 \\
UBP & 11.5 & 29.7 & 15.5 & 40.0 & 9.8 & 27.0 & 13.0 & 32.3 & 8.8 & 33.8 & 11.7 & 31.0 & 10.2 & 23.8 & 12.2 & 32.2 & \underline{15.5} & \underline{40.5} & 16.0 & 43.5 & 12.4 & 33.4 \\
NeuroBridge & \textbf{23.2} & \textbf{52.4} & \underline{21.2} & \underline{49.3} & \underline{13.2} & \underline{36.5} & 17.0 & \underline{45.3} & \underline{14.5} & \underline{37.7} & \textbf{25.0} & \textbf{55.0} & 15.3 & \underline{45.1} & \textbf{20.1} & \underline{44.9} & 13.7 & 36.5 & \underline{27.2} & \underline{56.3} & \underline{19.0} & \underline{45.9} \\
HyFI & 16.2 & 35.8 & 20.0 & 47.7 & 7.5 & 26.7 & \underline{18.8} & 41.3 & 9.7 & 27.5 & 15.5 & 33.8 & 11.0 & 34.5 & 13.2 & 30.3 & 13.3 & 38.8 & 25.3 & 55.7 & 15.1 & 37.2 \\
\textbf{SIMON} & \underline{20.0} & \underline{49.3} & \textbf{21.7} & \textbf{55.7} & \textbf{13.8} & \textbf{39.7} & \textbf{19.7} & \textbf{50.7} & \textbf{15.3} & \textbf{42.7} & \underline{23.0} & \underline{51.5} & \textbf{18.7} & \textbf{46.7} & \underline{17.8} & \textbf{48.2} & \textbf{18.7} & \textbf{47.7} & \textbf{27.3} & \textbf{66.8} & \textbf{19.6} & \textbf{49.9} \\
\bottomrule
\end{tabular}%
}
\end{table}

In the intra-subject setting, SIMON reaches an average Top-1 accuracy of \textbf{69.7\%} and a Top-5 accuracy of \textbf{92.9\%}, outperforming the previous state-of-the-art HyFI~\cite{jo2026hyfi}, which reports 68.2\% and 91.9\%, respectively. The improvement is consistent across most subjects, with particularly notable gains on Subject 4 and Subject 5, supporting the effectiveness of saliency-aware integration for capturing subject-specific neural patterns.

In the inter-subject setting, where the model is evaluated in a leave-one-subject-out manner, SIMON achieves \textbf{19.6\%} Top-1 and \textbf{49.9\%} Top-5 accuracy. This surpasses NeuroBridge~\cite{zhang2025neurobridge}, which achieves 19.0\% Top-1 and 45.9\% Top-5. The larger Top-5 margin (+4.0\%) suggests that SIMON learns more robust cross-subject representations and better bridges the domain gap across individuals.

\subsection{Validation of Geometric--Semantic Dissociation}
\label{sec:dissociation_validation}

For each test image, the semantic--geometric offset is quantified as the average spatial displacement between the image geometric center and the $K$ semantic centers derived from the saliency distribution. The resulting offsets are grouped into three intervals, $[0,33)$, $[33,66)$, and $[66,100)$, corresponding to mild, moderate, and severe dissociation. To assess the stability of the trend, the analysis is repeated over five random seeds, and the mean Top-1 gain of SIMON over a geometric-center baseline is reported together with the standard deviation.

\begin{figure}[htbp]
    \centering
    
    
    \begin{minipage}[t]{0.4\linewidth}
        \centering
        \vspace{0pt}
        \includegraphics[width=0.75\linewidth]{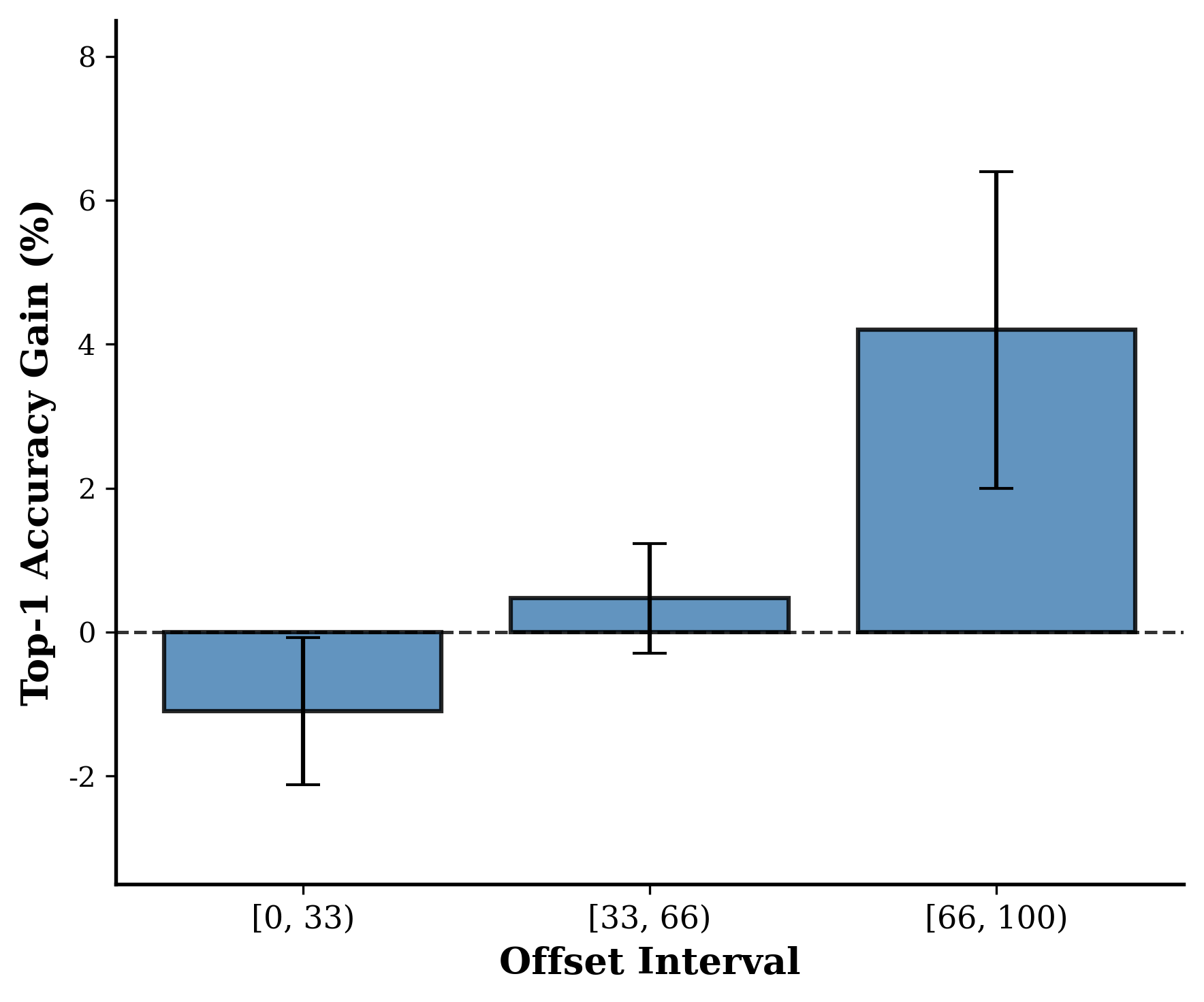}
    \end{minipage}%
    \hfill 
    \begin{minipage}[t]{0.55\linewidth}
        \centering
        \vspace{0pt}
        \resizebox{0.75\linewidth}{!}{%
            \begin{tabular}{lll cc}
            \toprule
            \multirow{2}{*}{\textbf{Saliency Extraction}} & \multirow{2}{*}{\textbf{Sampling}} & \multirow{2}{*}{\textbf{View}} & \multicolumn{2}{c}{\textbf{$\Delta$ Top-1 (\%)}} \\
            \cmidrule(lr){4-5}
            & & & \textbf{Intra} & \textbf{Inter} \\
            \midrule
            None & Random & Single & -1.5 & -0.7 \\
                 & Random & Multi  & -1.3 & -0.2 \\
            \midrule
            \multirow{4}{*}{Foreground} & Random & Single & -1.7 & -0.4 \\
                                        & Random & Multi  & -1.8 & -0.8 \\
                                        & Saliency-aware & Single & -1.1 & -0.2 \\
                                        & Saliency-aware & Multi  & -0.6 & -0.2 \\
            \midrule
            \multirow{4}{*}{\begin{tabular}{@{}l@{}}Foreground \\ + Saliency\end{tabular}} 
                                        & Random & Single & -1.8 & -0.4 \\
                                        & Random & Multi  & -1.2 & -0.5 \\
                                        & Saliency-aware & Single & -1.5 & -0.1 \\
                                        & Saliency-aware & Multi  & \textbf{0.0} & \textbf{0.0} \\
            \bottomrule
            \end{tabular}%
        }
    \end{minipage}
    
    \vspace{0.3cm} 
    
    
    \begin{minipage}[t]{0.4\linewidth}
        \captionsetup{justification=raggedright, font=small} 
        \captionof{figure}{Dissociation-conditioned analysis of the Top-1 intra-subject retrieval gain.}
        \label{fig:dissociation_conditioned}
    \end{minipage}%
    \hfill 
    \begin{minipage}[t]{0.55\linewidth}
        \captionsetup{justification=raggedright, font=small} 
        \captionof{table}{Ablation analysis of saliency extraction, sampling strategy, and view generation.}
        \label{tab:three_factor_ablation_full_text}
    \end{minipage}

    \vspace{-0.5cm} 
    
\end{figure}

Figure~\ref{fig:dissociation_conditioned} shows a monotonic increase in performance gain as the semantic--geometric offset increases. While gains in the lower intervals ($[0,66)$) are statistically non-significant, the highest interval $[66,100)$ shows a gain of $4.20\%\pm2.20\%$ ($p = 0.013$, one-sample $t$-test). A one-way ANOVA indicates that the offset magnitude has a significant effect on the performance gain ($p < 0.001$).

The pattern is consistent with the role of Geometric--Semantic Dissociation in center-fixed viewing. The largest improvement appears in the regime where the geometric center is least likely to coincide with semantically informative regions. Performance gains are therefore concentrated in samples with larger semantic--geometric mismatch, rather than being distributed uniformly across the test set.

\subsection{Ablation Studies}
\label{sec:ablation}

\textbf{Ablation study on saliency-aware multi-view encoding.}\label{sec:three_factor_ablation} Table~\ref{tab:three_factor_ablation_full_text} examines the effects of saliency extraction, sampling strategy, and view generation. Compared to the full configuration, multi-view generation alone leads to a performance decrease of 1.3\% in the intra-subject setting and 0.2\% in the inter-subject setting. Applying foreground extraction or joint guidance (foreground + saliency) under random sampling results in performance reductions of up to 1.8\% in the intra-subject setting.

Saliency-aware sampling limits these drops by selecting more informative regions. The optimal performance is obtained through the combination of multi-view generation, saliency-aware sampling, and joint foreground and saliency guidance. Removing these components decreases accuracy by 1.5\% and 0.7\% in the intra-subject and inter-subject settings, respectively.

The effectiveness of multi-view encoding depends on candidate region selection. Additional views or guidance priors alone do not sustain optimal performance. Its benefit emerges when combined with saliency-aware selection.

\begin{figure}[htbp]
    \centering
    \begin{subfigure}[b]{0.3\textwidth}
        \centering
        \includegraphics[width=\linewidth]{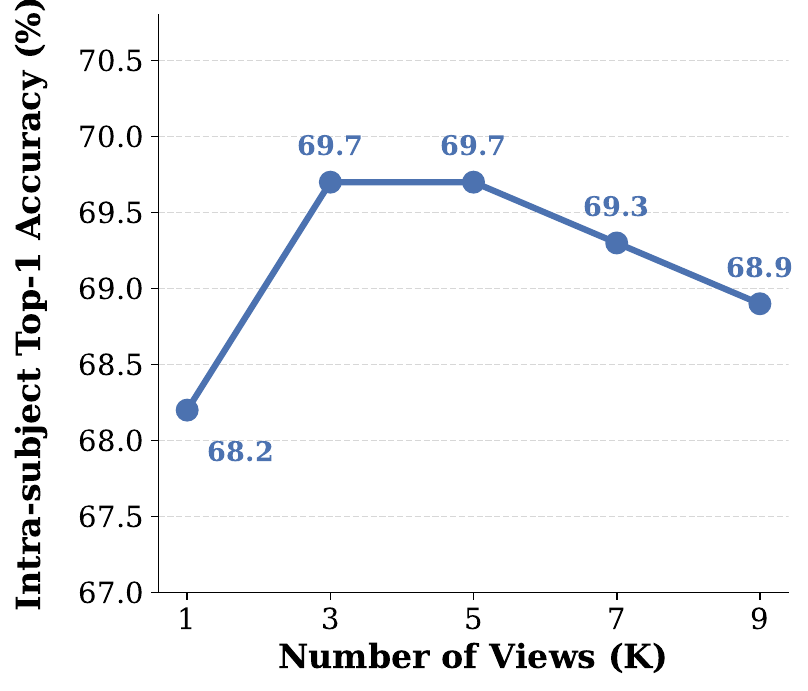}
    \end{subfigure}
    \hspace{0.05\textwidth} 
    \begin{subfigure}[b]{0.3\textwidth}
        \centering
        \includegraphics[width=\linewidth]{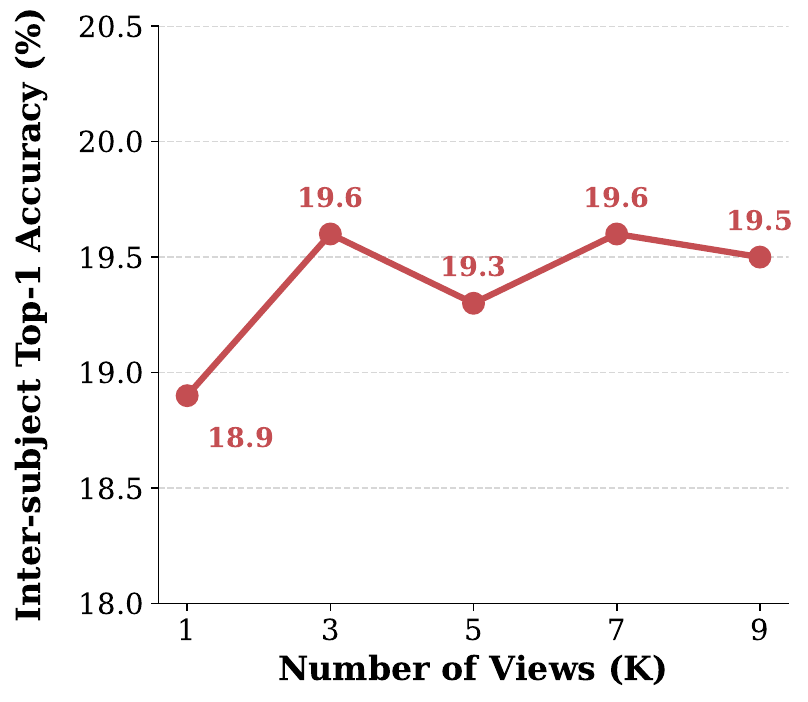}
    \end{subfigure}
    \caption{\textbf{Sensitivity of Top-1 and Top-5 retrieval accuracy to the sampling granularity .}}
    \label{fig:k_sensitivity}
\end{figure}

\textbf{Sensitivity to Saliency-Aware Sampling Granularity.}\label{sec:ablation_k} To validate our bio-inspired aggregation strategy, we analyze retrieval performance as a function of sampling granularity \(K\), which denotes the number of simulated fixation centers. Figure~\ref{fig:k_sensitivity} shows the Top-1 accuracy trends under both intra- and inter-subject settings. 

The single-view baseline (\(K=1\)) performs relatively poorly, reaching 68.2\% in the intra-subject setting and 18.9\% in the inter-subject setting, suggesting that a single foveal crop cannot capture the full semantics of the visual stimulus. As \(K\) increases, performance improves in both settings, indicating that aggregating multiple attended regions better captures visual evidence. The best performance is achieved at \(K=3\), with Top-1 accuracies of \textbf{69.7\%} and \textbf{19.6\%} in the intra- and inter-subject settings, respectively.

As $K$ increases further, the trends diverge. Intra-subject performance declines after $K=5$, suggesting excessive views introduce low-saliency noise. In contrast, inter-subject performance remains stable at higher granularities, showing greater tolerance for diverse visual evidence. Overall, $K=3$ remains the optimal choice for both settings.

\subsection{Analysis}
\label{sec:analysis}

\begin{figure*}[htbp]
    \centering
    \begin{subfigure}[b]{0.27\textwidth}
        \centering
        \raisebox{0.5cm}{\includegraphics[width=\linewidth]{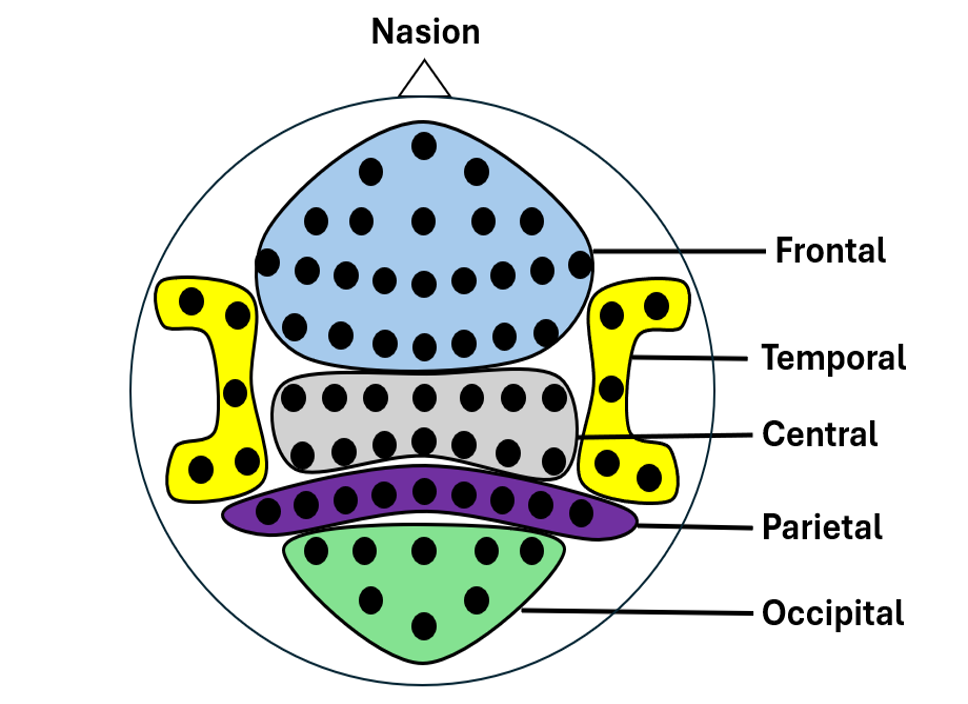}}
        \small (c)
        \label{fig:channel-grouping}
    \end{subfigure}
    \hfill
    \begin{subfigure}[b]{0.33\textwidth}
        \centering
        \includegraphics[width=\linewidth]{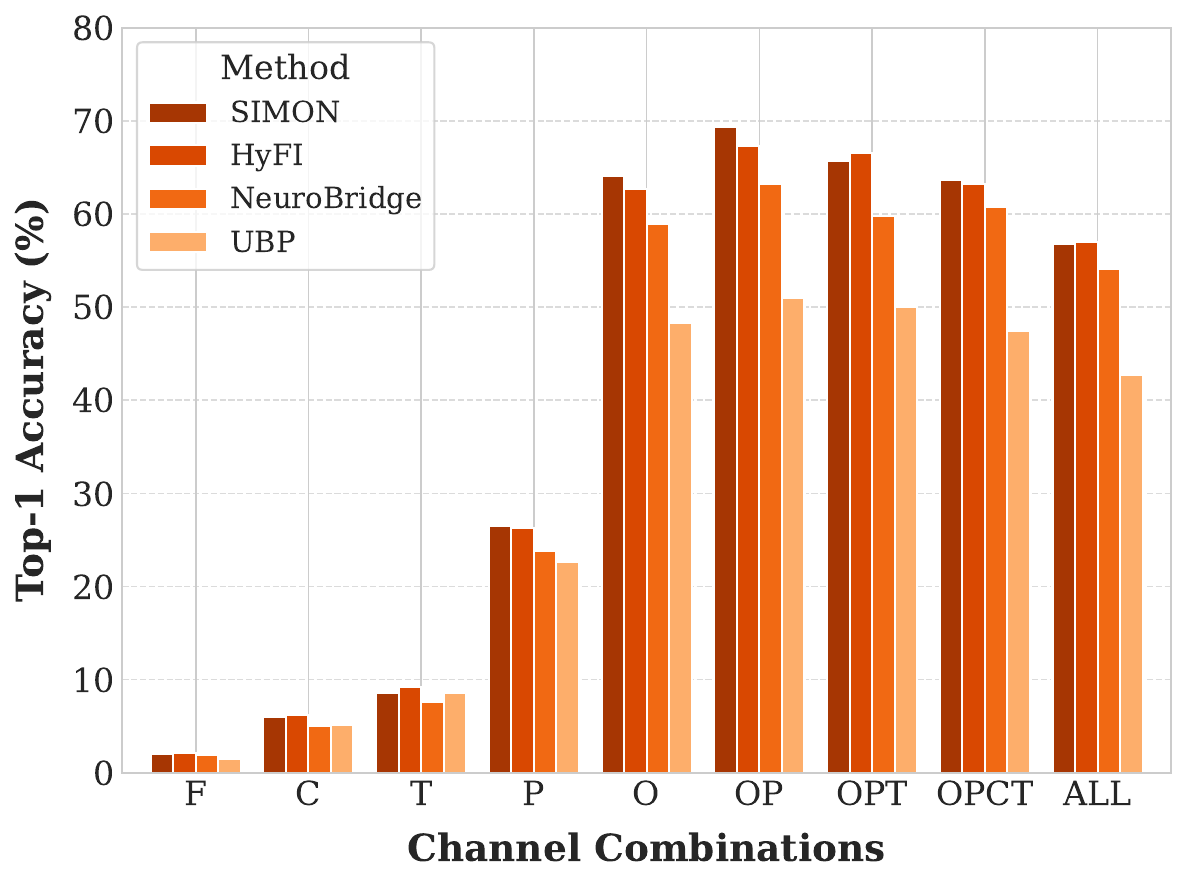}
        \small (d)
        \label{fig:intra_perf}
    \end{subfigure}
    \hfill
    \begin{subfigure}[b]{0.33\textwidth}
        \centering
        \includegraphics[width=\linewidth]{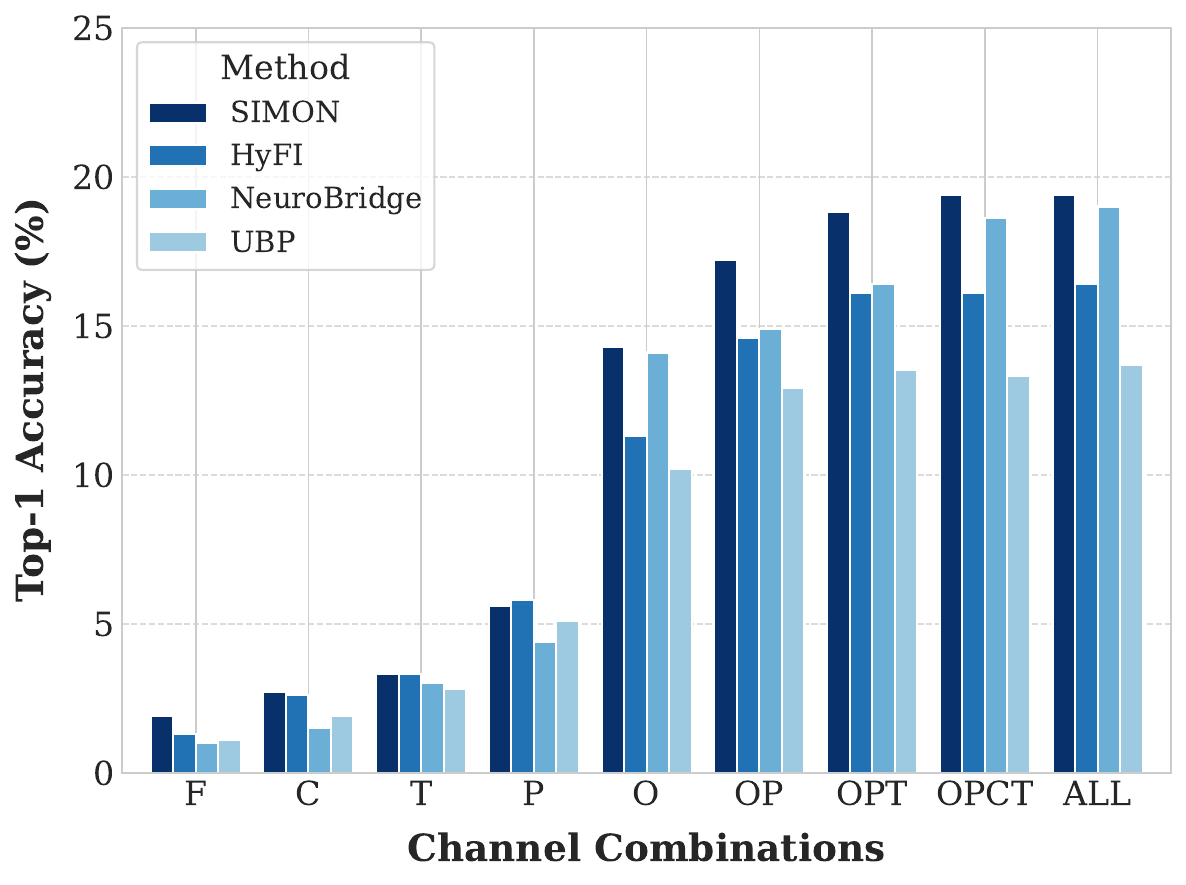}
        \small (e)
        \label{fig:inter_perf}
    \end{subfigure}
    \caption{
        \textbf{Impact of channel combinations on retrieval performance and channel grouping visualization.}
        (\textbf{c}) Channel grouping. The EEG electrodes (black dots, 10--10 montage) are grouped into five regions: Frontal (F), Central (C), Temporal (T), Parietal (P), and Occipital (O).
        (\textbf{d}) Impact of channel combinations on intra-subject retrieval performance.
        (\textbf{e}) Impact of channel combinations on inter-subject retrieval performance across different methods.
    }
    \label{fig:channel_effect_comparison}
\end{figure*}

\textbf{Impact of Channel Topology.}\label{sec:channel_ablation} Prior work such as UBP~\cite{wu2025bridging} suggests that posterior channels, particularly the Occipital and Parietal (OP) regions, are sufficient for effective decoding. The present analysis indicates that this pattern depends on the evaluation setting. Figure~\ref{fig:channel_effect_comparison} shows that intra-subject retrieval generally favors posterior-dominant configurations, where OP yields the strongest performance and the inclusion of anterior channels tends to reduce accuracy. In inter-subject retrieval, the trend is reversed: broader channel configurations consistently improve performance across methods. For example, UBP improves from 12.9\% (OP) to 13.7\% (ALL), HyFI from 14.6\% to 16.4\%, and NeuroBridge from 14.9\% to 19.0\%. This contrast indicates that posterior channels capture strong subject-specific visual signals, whereas broader spatial coverage provides more invariant information for cross-subject generalization.

\begin{figure}[htbp]
    \centering
    
    \begin{minipage}[t]{0.48\linewidth}
        \centering
        \includegraphics[width=\linewidth]{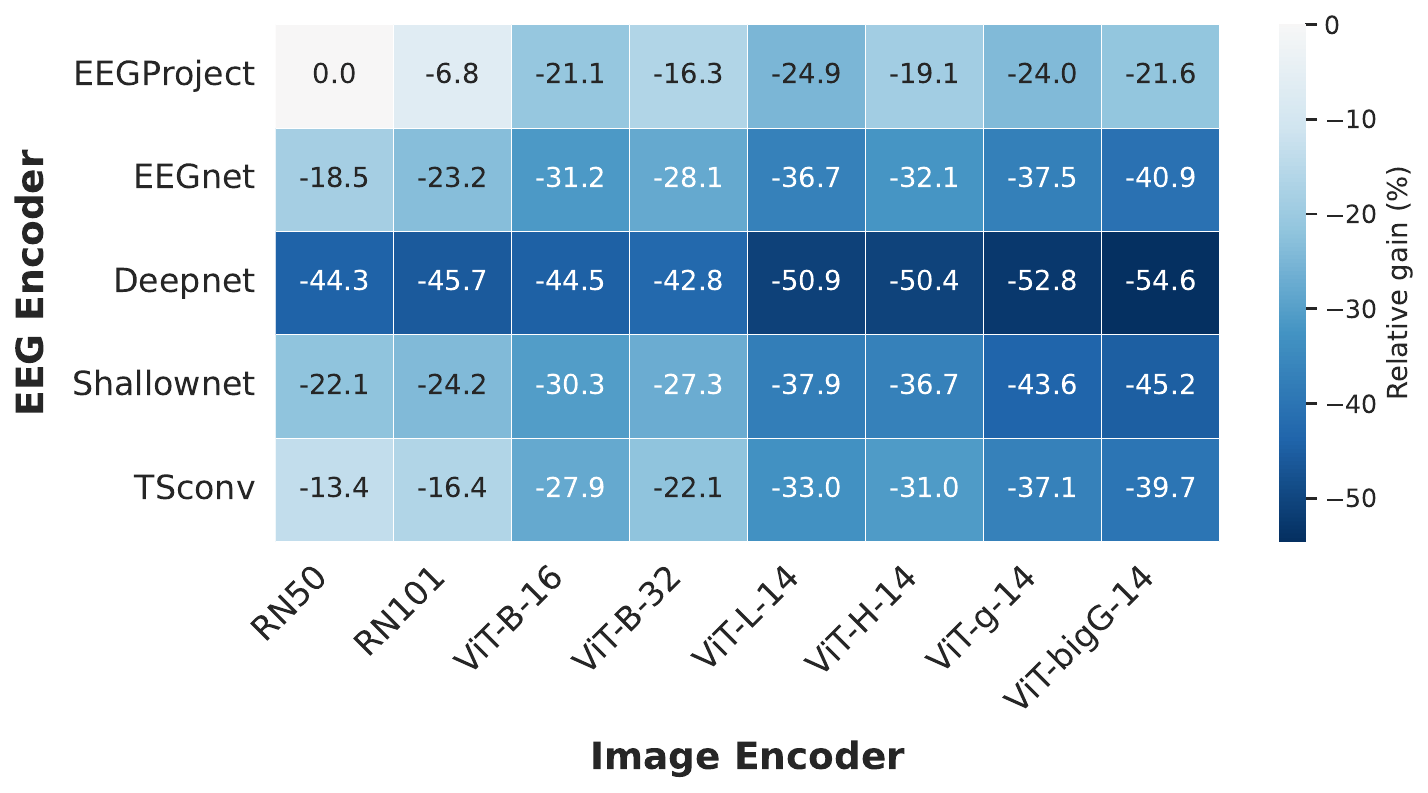}
        
        \vspace{0.1cm} 
        \caption{Relative Top-1 accuracy gain of SIMON across different EEG and image encoders.}
        \label{fig:encoder_heatmap_top1}
    \end{minipage}
    \hfill
    \begin{minipage}[t]{0.48\linewidth}
        \centering
        \includegraphics[width=\linewidth]{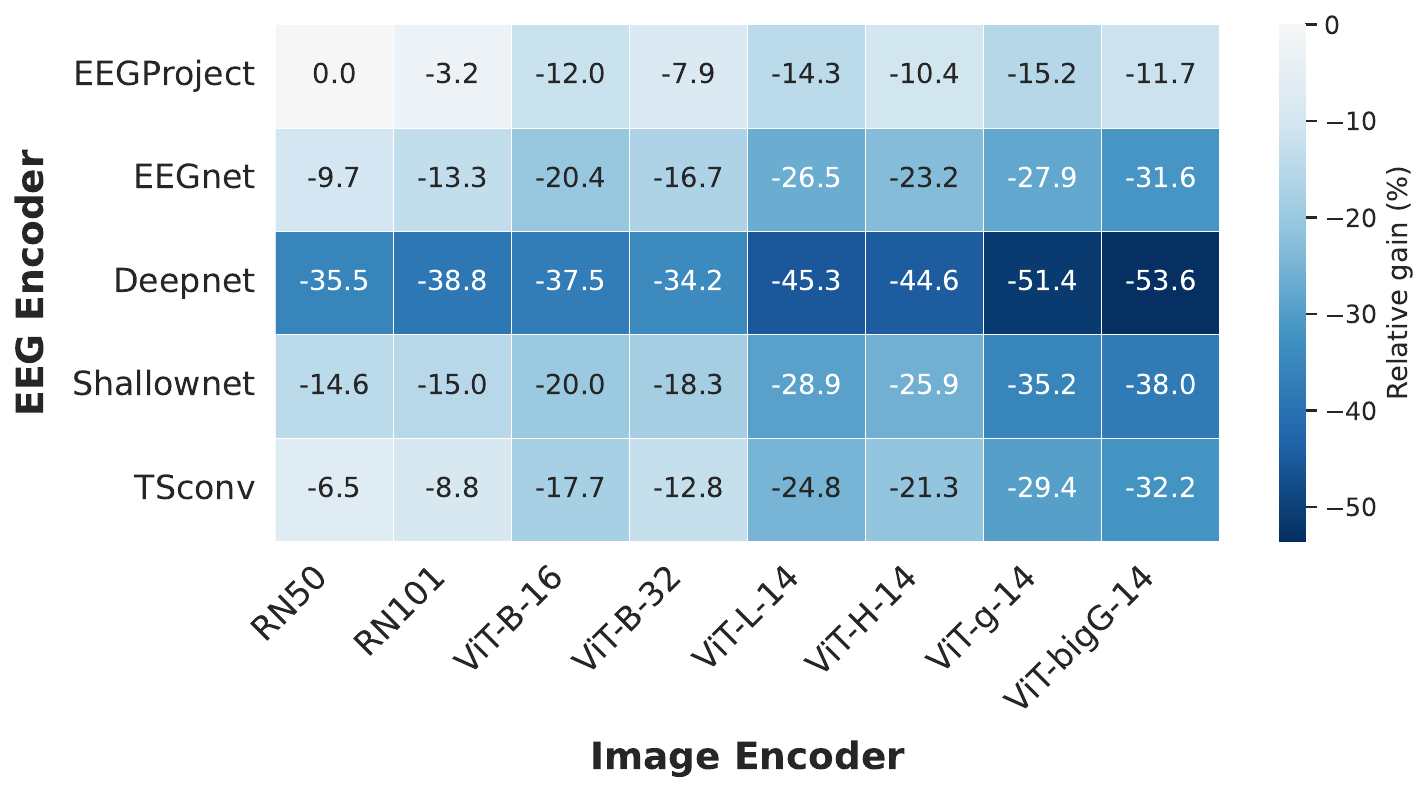}
        
        \vspace{0.1cm}
        \caption{Relative Top-5 accuracy gain of SIMON across different EEG and image encoders.}
        \label{fig:encoder_heatmap_top5}
    \end{minipage}
    
\end{figure}

\textbf{Effect of Vision and Brain Encoders.}\label{sec:encoder_analysis} To examine how visual representations interact with neural decoding capacity, we evaluate retrieval performance across different combinations of vision and brain encoders. Figure~\ref{fig:encoder_heatmap_top1} and \ref{fig:encoder_heatmap_top5} reports the relative Top-1 and Top-5 accuracy variations obtained by pairing multiple EEG encoders with ResNet- and Vision Transformer-based visual backbones. Across these combinations, the brain encoder has a clear influence on retrieval accuracy. EEGProject consistently yields the strongest relative performance, while replacing it with alternative EEG encoders results in lower accuracy across nearly all visual backbones. The degradation is most pronounced for Deepnet, whereas the other EEG encoders show smaller but consistent reductions. These results suggest that the quality of neural feature extraction plays a central role in determining cross-modal retrieval performance.

The choice of vision encoder affects the absolute retrieval scores, but it does not substantially change the relative ordering among EEG encoders. This pattern is observed for both CNN-based and transformer-based visual representations, indicating that stronger or larger visual backbones do not necessarily compensate for weaker neural representations. The persistent gap between EEGProject and the other EEG encoders suggests that, in this setting, retrieval performance is more sensitive to the robustness and alignment quality of the EEG representation than to the specific family or scale of the visual embedding space.

\subsection{Qualitative Results}
\label{sec:visualization}

\begin{figure}[htbp]
    \centering
    \setlength{\tabcolsep}{1pt} 
    \renewcommand{\arraystretch}{0}
    \setlength\dashlinedash{3pt}
    \setlength\dashlinegap{2pt}
    \newcommand{\imgwidth}{0.08\linewidth} 

    \begin{tabular}{c : ccc : ccc}
        \multicolumn{1}{c}{\scriptsize \textbf{Target}} &
        \multicolumn{3}{c}{\scriptsize \textbf{SIMON}} &
        \multicolumn{3}{c}{\scriptsize \textbf{HyFI (prev. SOTA)}} \\ \noalign{\vspace{2pt}}

        \multicolumn{1}{c}{} &
        \tiny Top-1 & \tiny Top-2 & \multicolumn{1}{c}{\tiny Top-3} &
        \tiny Top-1 & \tiny Top-2 & \tiny Top-3 \\ \noalign{\vspace{2pt}}

        \includegraphics[width=\imgwidth]{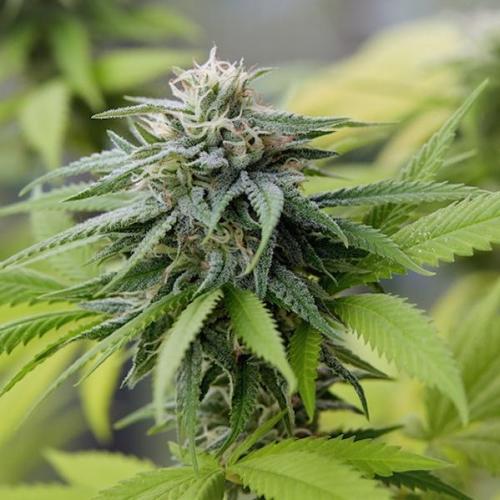} &
        \includegraphics[width=\imgwidth]{thingsimg/marijuana_112.jpg} &
        \includegraphics[width=\imgwidth]{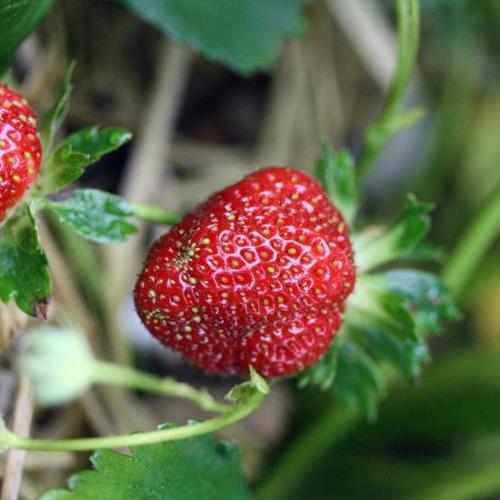} &
        \includegraphics[width=\imgwidth]{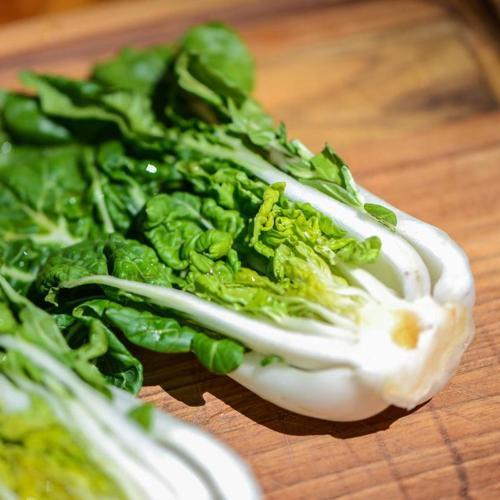} &
        \includegraphics[width=\imgwidth]{thingsimg/bok_choy_18.jpg} &
        \includegraphics[width=\imgwidth]{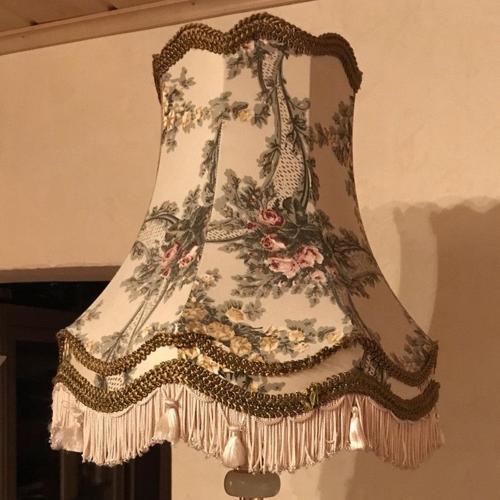} &
        \includegraphics[width=\imgwidth]{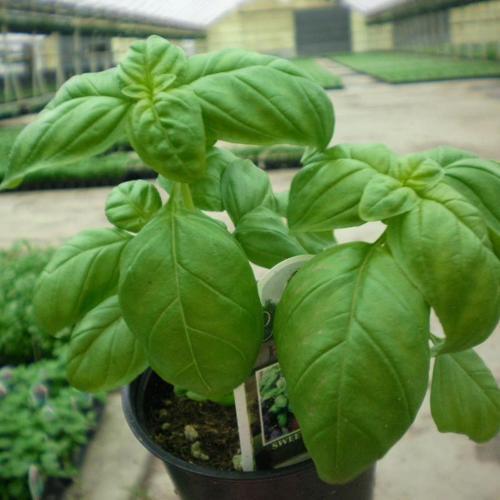} \\
        \noalign{\vspace{1pt}}

        \includegraphics[width=\imgwidth]{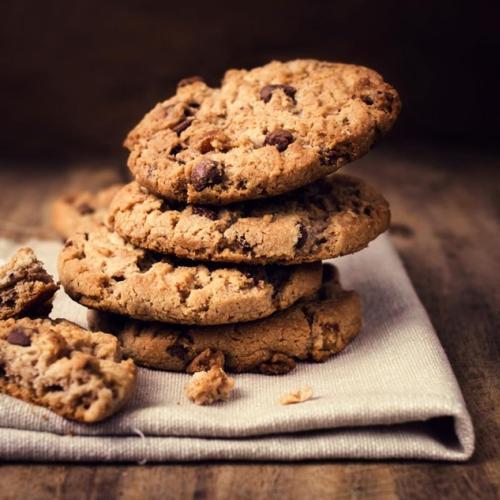} &
        \includegraphics[width=\imgwidth]{thingsimg/cookie_50.jpg} &
        \includegraphics[width=\imgwidth]{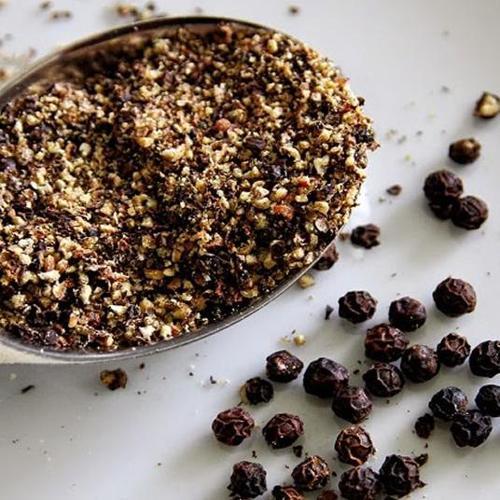} &
        \includegraphics[width=\imgwidth]{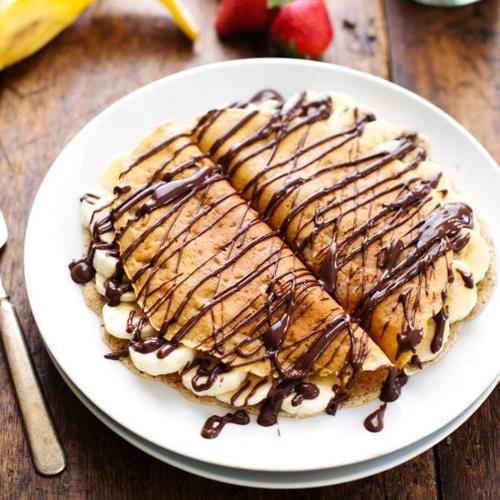} &
        \includegraphics[width=\imgwidth]{thingsimg/crepe_55.jpg} &
        \includegraphics[width=\imgwidth]{thingsimg/lampshade_107.jpg} &
        \includegraphics[width=\imgwidth]{thingsimg/pepper1_132.jpg} \\
        \noalign{\vspace{1pt}}

        \includegraphics[width=\imgwidth]{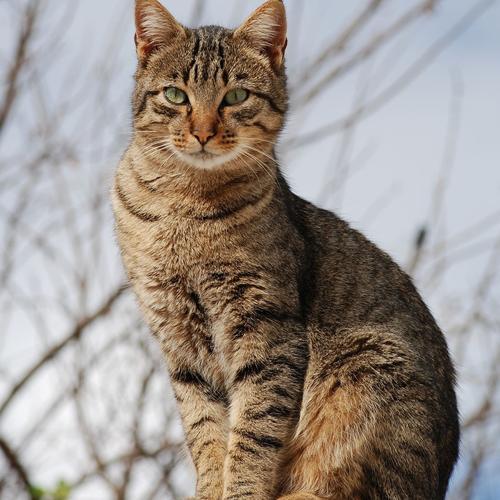} &
        \includegraphics[width=\imgwidth]{thingsimg/cat_33.jpg} &
        \includegraphics[width=\imgwidth]{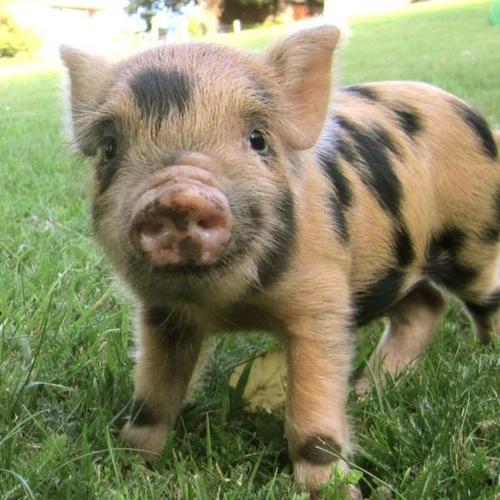} &
        \includegraphics[width=\imgwidth]{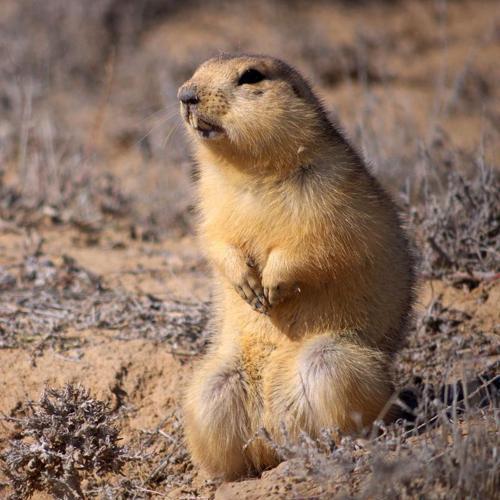} &
        \includegraphics[width=\imgwidth]{thingsimg/gopher_87.jpg} &
        \includegraphics[width=\imgwidth]{thingsimg/piglet_137.jpg} &
        \includegraphics[width=\imgwidth]{thingsimg/cat_33.jpg} \\
    \end{tabular}

    \vspace{2pt} 
    \caption{Visualization of Top-3 retrieved images on the THINGS-EEG dataset (SIMON vs HyFI).}
    \label{fig:comparison}
\end{figure}

Figure~\ref{fig:comparison} shows the Top-3 retrieval results and illustrates how the methods differ in the visual evidence they rely on. Models driven mainly by global statistics, such as object shape, layout, or background, can overemphasize coarse similarity and produce semantically weaker matches. In the cookie example, this bias leads to confusion with a \textit{lampshade}, which has a similar stacked silhouette but different object semantics. SIMON preserves more local discriminative cues, such as granular surface texture, and retrieves semantically closer items such as \textit{pepper}, even when the exact target is not ranked first.

A similar pattern appears when the target and distractors share comparable global appearance. In the \textit{Cat} example, object-specific regions such as facial features provide useful evidence for distinguishing the target from related alternatives. The saliency-aware views reduce the influence of background or coarse layout similarity and support more accurate identification. In the \textit{Plant} case, the results further suggest that local details help separate visually similar instances within the same broad category, indicating that the gains are associated with both category-level semantics and instance-level discrimination.

\section{Limitations and Future Work}
\label{sec:limitations}

Although the framework addresses the geometric-semantic dissociation between visual features and neural responses , the multi-stage encoding process introduces additional computational requirements. The combination of saliency estimation and saliency-aware sampling  adds a processing latency of approximately 0.87 seconds per image. This overhead is acceptable for offline retrieval tasks but serves as a constraint for deployment in real-time systems where rapid feedback loops are necessary. Future studies will focus on methods to increase the efficiency of the visual pipeline by implementing algorithmic parallelization or adopting lightweight saliency backbones to reduce total inference time while maintaining semantic precision.

\section{Conclusion}
\label{sec:conclusion}

We address Geometric–Semantic Dissociation in EEG-to-image retrieval, where center-fixed foveation conflicts with content-driven attention and misaligns visual features with EEG. We propose \textbf{SIMON}, a saliency-aware multi-view framework that selects fixation centers via Saliency-Aware Sampling and generates foveated views emphasizing semantic foregrounds while suppressing background clutter. We quantify this dissociation on THINGS images and show it is widespread. On THINGS-EEG, SIMON achieves state-of-the-art performance with consistent gains in both intra- and inter-subject settings, particularly improving cross-subject generalization. Analyses across sampling density, channel topology, and encoder backbones further support the robustness and generality of saliency-aware multi-view integration. Beyond retrieval performance, this line of research may support future non-invasive BCI and assistive communication studies, but it also requires careful attention to privacy, consent, and potential misuse of neural decoding technologies.

\bibliographystyle{plainnat}
\bibliography{references}

\newpage
\appendix

\section{Experiment Configuration}

\subsection{Datasets and Preprocessing}

\textbf{Dataset.} We evaluate our method on the THINGS-EEG dataset~\cite{gifford2022large}, a large-scale benchmark collected using a 64-channel EASYCAP system under the Rapid Serial Visual Presentation (RSVP) paradigm. The dataset contains EEG recordings from 10 subjects. The training partition consists of 1,654 object concepts, where each concept is represented by 10 unique images presented 4 times, yielding a total of 16,540 training instances per subject. Conversely, the test partition comprises 200 concepts with one image per concept, repeated 80 times to ensure robust evaluation.

\textbf{Preprocessing.} Following the standard methodology outlined by \cite{song2024decoding, wu2025bridging}, we processed the raw EEG signals by extracting the [0, 1000] ms time window relative to the stimulus onset and applying a baseline correction based on the mean activity of the preceding 200 ms. The data were resampled to 250 Hz, and trials corresponding to the same image were averaged to mitigate noise.

\subsection{Implementation Details}

\textbf{Experimental Setup.} We implemented our method using Python 3.10 and PyTorch 2.4.0 on a workstation equipped with an Intel Core i7-14700 CPU, 128 GB of RAM, and a single NVIDIA GeForce RTX 4090 GPU (24 GB). For training, we utilize the AdamW optimizer~\cite{loshchilov2019decoupled} with a fixed weight decay of $1\times10^{-4}$, a batch size of 1,024, and a total of 50 epochs. The learning rate is set to $3\times10^{-4}$ for intra-subject experiments and $3\times10^{-5}$ for inter-subject experiments to ensure stable convergence.

\textbf{Brain Encoder Architectures.} To optimally capture neural representations across different tasks, we employ distinct backbone architectures tailored to the specific challenges of each setting. For the \textbf{intra-subject} task, we utilize the EEGProject architecture~\cite{wu2025bridging} with a focused selection of 17 channels covering the occipital and parietal regions. In contrast, for the \textbf{inter-subject} task, we adopt the TSconv backbone~\cite{song2024decoding} and utilize all 63 channels (excluding the Fz reference) to capture a comprehensive, whole-brain receptive field. This configuration aligns with our findings that broader spatial integration is essential for robust cross-subject generalization.

\textbf{Visual Feature Extraction.} To extract high-level visual embeddings, we employ the OpenCLIP~\cite{cherti2023reproducible} implementation of ResNet-50~\cite{he2016deep} initialized with OpenAI pre-trained weights. All images are resized to $224 \times 224$ pixels. To facilitate the proposed Saliency-Aware Multi-View Visual Encoding, we utilize BiRefNet~\cite{zheng2024birefnet} for background removal and SUM~\cite{hosseini2025sum} for saliency detection. Based on these maps, we generate $K$ distinct foveated views for each image using a Gaussian blur pyramid, configuring $K=3$ for both intra-subject and inter-subject tasks. Unless otherwise specified, we set the foreground threshold to $\tau=0.5$.

\textbf{Model Variants.} To evaluate the robustness of our framework, we benchmarked various backbone architectures. For the brain encoder, we compared our selected models against standard baselines including EEGNet~\cite{lawhern2018eegnet}, Deepnet~\cite{schirrmeister2017deep}, and Shallownet~\cite{schirrmeister2017deep}. For the vision encoder, we evaluated a wide range of architectures: ResNet (RN50, RN101) and Vision Transformers (ViT-B-32, ViT-B-16, ViT-L-14, ViT-H-14, ViT-bigG-14, ViT-g-14).

\subsection{Licenses of Existing Assets}
Our research builds upon several publicly available datasets and pre-trained models. We properly cite the original creators and strictly adhere to their terms of use:
\begin{itemize}
    \item \textbf{THINGS-EEG \cite{gifford2022large}:} The dataset is publicly available and distributed under the Creative Commons Attribution 4.0 International (\textbf{CC BY 4.0}) License.
    \item \textbf{BiRefNet \cite{zheng2024birefnet}:} The official PyTorch implementation and pre-trained weights are utilized under the \textbf{MIT License}.
    \item \textbf{SUM \cite{hosseini2025sum}:} The official codebase provided by the authors is utilized under the \textbf{MIT License}.
\end{itemize}

\section{Additional Experimental Results}

\subsection{Detailed Analysis of the Sampling Factor}
\label{app:ablation_sampling}

\begin{figure}[htbp]
\centering
\includegraphics[width=0.4\textwidth]{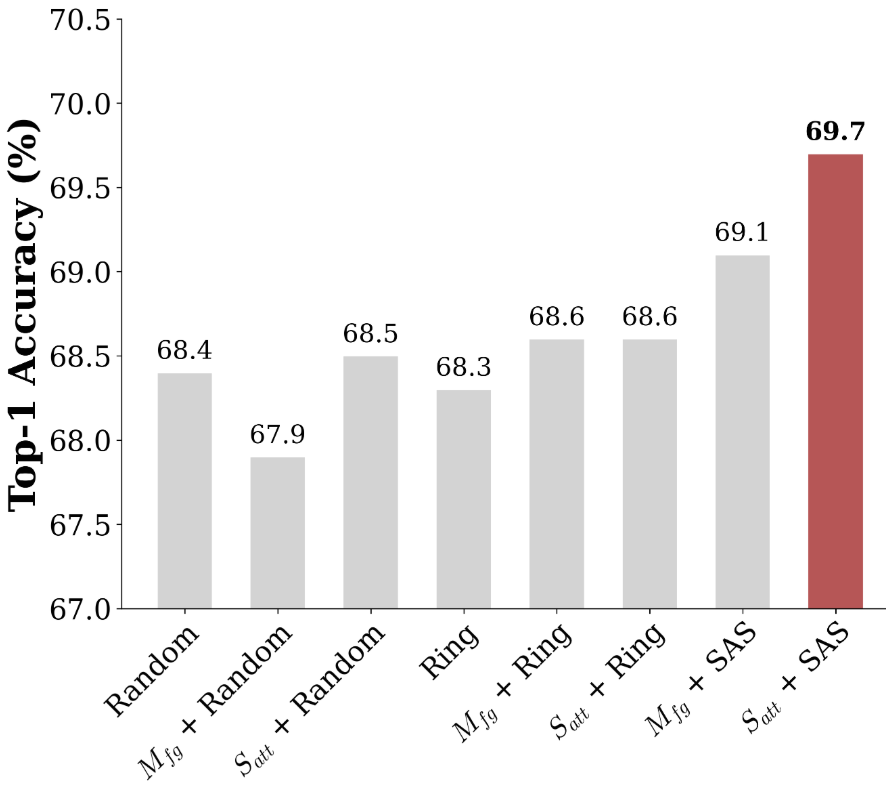}
\caption{Comparison of visual sampling configurations.}
\label{fig:ablation_sampling}
\end{figure}

This section provides a more detailed analysis of the sampling strategies summarized in Figure~\ref{fig:ablation_sampling}. Building on the ablation study in the main text, we further examine how fixation centers are selected under different levels of visual guidance. This analysis focuses on the sampling strategy itself, while keeping the roles of saliency extraction and view generation consistent with the main experimental design.

\textbf{Sampling Strategy under Different Saliency Extraction Settings.}
We compare sampling strategies under three saliency extraction settings: None, Foreground, and Foreground + Saliency. When saliency extraction is set to None, fixation centers are selected from the full image, and the initial center is fixed at the geometric center. This setting does not suppress background regions and therefore may sample visually irrelevant areas.

When saliency extraction is set to Foreground, the candidate region is restricted by the foreground mask $M_{fg}$. This reduces background interference, but the sampling process remains largely geometry-driven because the foreground mask mainly indicates object support rather than visual importance. As a result, this setting improves spatial validity but may still assign views to object regions that are distant yet semantically weak.

When saliency extraction is set to Foreground + Saliency, the candidate region is constrained by the foreground mask and further weighted by the saliency map. In this setting, Saliency-Aware Sampling selects fixation centers according to
\[
J(p) = D(p) \cdot S_{att}(p)^\gamma, \quad p \in \Omega,
\]
where $\Omega = \{p \mid M_{fg}(p) > \tau\}$ and $D(p)$ denotes the distance from previously selected centers. This strategy encourages the selected views to be both spatially diverse and visually informative.

\textbf{Comparison of Sampling Rules.}
Within these saliency extraction settings, we compare three sampling rules: random sampling, ring sampling, and Saliency-Aware Sampling. Random sampling provides stochastic spatial coverage but does not explicitly optimize the distribution of fixation centers. Ring sampling provides a deterministic geometric pattern around the initial center, but it cannot adapt to irregular object shapes or non-uniform semantic importance. In contrast, Saliency-Aware Sampling explicitly combines spatial diversity with saliency weighting, allowing the model to allocate views to informative object regions rather than uniformly or geometrically distributed locations.

\textbf{Analysis of Results.}
As shown in Figure~\ref{fig:ablation_sampling}, sampling without saliency extraction performs worse because the selected views may include background clutter. Foreground-based sampling improves over the no-extraction setting by restricting candidate points to object regions, but its gains are limited because it mainly follows object geometry. The best performance is achieved when Foreground + Saliency is combined with Saliency-Aware Sampling, which corresponds to the full SIMON configuration. These results support the main ablation conclusion that multi-view generation is most effective when additional views are selected from semantically informative foreground regions.

\subsection{Effect of different EEG Channel Groups}
\label{app:channel_effect}

While the main text focuses on Top-1 accuracy to highlight the optimal topology for precise retrieval, \Cref{tab:intra_ch} and \Cref{tab:inter_ch} provide a more comprehensive evaluation of the channel partitions shown in \Cref{fig:channel-grouping} by including Top-5 accuracy. This metric offers insight into the model's ability to localize the correct target within a small candidate pool, effectively measuring the robustness of the learned representation against noise.

\textbf{Noise Tolerance in Intra-Subject Settings.}
Comparing Top-1 and Top-5 metrics in \Cref{tab:intra_ch} reveals the nature of the interference caused by anterior channels (T, C, F). For SIMON, adding these channels causes a sharp 12.9\% drop in Top-1 accuracy (69.7\% $\rightarrow$ 56.8\%). However, the Top-5 accuracy degrades less severely, dropping by only 5.2\% (92.9\% $\rightarrow$ 87.7\%). A similar pattern is observed for NeuroBridge, where Top-1 accuracy falls notably by 9.1\% (63.2\% $\rightarrow$ 54.1\%) while Top-5 accuracy only decreases by 4.2\% (90.2\% $\rightarrow$ 86.0\%). This discrepancy suggests that while irrelevant signals from non-visual areas confuse the model's precise discrimination (lowering Top-1), the core semantic information remains partially accessible, keeping the correct answer within the top candidates. Thus, in intra-subject tasks, spatial focus is critical for \textit{precision}, but the system retains some \textit{recall} capacity even with noisy inputs.

\textbf{Expanding the "Retrieval Neighborhood" in Inter-Subject Generalization.}
In the inter-subject domain (\Cref{tab:inter_ch}), the inclusion of Top-5 data reinforces the necessity of broad spatial integration. The gains observed in Top-1 are amplified in Top-5 performance. 
For instance, SIMON's Top-5 accuracy surges from 44.1\% (OP) to 49.9\% (ALL), and NeuroBridge exhibits a similarly substantial improvement from 39.8\% (OP) to 45.9\% (ALL). This indicates that nearly half of the time, the correct image is retrieved within the first five predictions when utilizing the full brain topology. 
Crucially, unlike the intra-subject case where the "ALL" configuration hurts performance, here the Top-5 metric continues to climb or stabilize for all evaluated methods. This implies that the additional channels (T, C) do not merely add noise but provide critical "anchoring" features that may not always be strong enough to force a Top-1 hit, but are sufficient to pull the correct target into the top-ranked cluster, significantly improving the retrieval system's utility.

\begin{table}[htbp]
\centering
\renewcommand{\arraystretch}{0.85} 
\caption{Intra-subject classification results on THINGS-EEG across different EEG channel combinations.}
\label{tab:intra_ch}
\resizebox{0.7\textwidth}{!}{%
\begin{tabular}{l c c c c c c c c}
\toprule
\multirow{2}{*}{\textbf{Channel}} 
& \multicolumn{2}{c}{\textbf{UBP}} 
& \multicolumn{2}{c}{\textbf{NeuroBridge}} 
& \multicolumn{2}{c}{\textbf{HyFI}} 
& \multicolumn{2}{c}{\textbf{SIMON}} \\
\cmidrule(lr){2-3} \cmidrule(lr){4-5} \cmidrule(lr){6-7} \cmidrule(lr){8-9}
 & Top-1 & Top-5 & Top-1 & Top-5 & Top-1 & Top-5 & Top-1 & Top-5 \\
\midrule
\textbf{F}    & 1.5 & 7.3  & 1.9 & 8.0  & 2.1 & 7.7  & 2.0 & 8.9  \\
\textbf{C}    & 5.1 & 17.5 & 5.0 & 16.8 & 6.2 & 19.4 & 6.0 & 18.9 \\
\textbf{T}    & 8.6 & 26.3 & 7.6 & 24.1 & 9.2 & 27.5 & 8.6 & 28.0 \\
\textbf{P}    & 22.6& 49.7 & 23.8& 51.9 & 26.3& 54.5 & 26.5& 57.3 \\
\textbf{O}    & 48.3& 78.4 & 58.9& 86.3 & 62.7& 89.7 & 64.1& 90.0 \\
\textbf{OP}   & 51.0& 79.6 & 63.2& 90.2 & 67.3& 92.0 & 69.7& 92.9 \\
\textbf{OPT}  & 50.0& 79.5 & 59.8& 89.3 & 66.5& 91.6 & 65.7& 92.2 \\
\textbf{OPCT} & 47.4& 78.0 & 60.8& 88.5 & 63.2& 90.8 & 63.6& 91.1 \\
\textbf{ALL}  & 42.7& 73.8 & 54.1& 86.0 & 57.0& 87.3 & 56.8& 87.7 \\
\bottomrule
\end{tabular}%
}

\vspace{0.4cm} 

\caption{Inter-subject classification results on THINGS-EEG across different EEG channel combinations.}
\label{tab:inter_ch}
\resizebox{0.7\textwidth}{!}{%
\begin{tabular}{l c c c c c c c c}
\toprule
\multirow{2}{*}{\textbf{Channel}} 
& \multicolumn{2}{c}{\textbf{UBP}} 
& \multicolumn{2}{c}{\textbf{NeuroBridge}} 
& \multicolumn{2}{c}{\textbf{HyFI}} 
& \multicolumn{2}{c}{\textbf{SIMON}} \\
\cmidrule(lr){2-3} \cmidrule(lr){4-5} \cmidrule(lr){6-7} \cmidrule(lr){8-9}
 & Top-1 & Top-5 & Top-1 & Top-5 & Top-1 & Top-5 & Top-1 & Top-5 \\
\midrule
\textbf{F}    & 1.1 & 5.0  & 1.0 & 4.3  & 1.3 & 5.0  & 1.9 & 6.5  \\
\textbf{C}    & 1.9 & 8.6  & 1.5 & 6.4  & 2.6 & 10.1 & 2.7 & 10.0 \\
\textbf{T}    & 2.8 & 12.6 & 3.0 & 10.9 & 3.3 & 13.8 & 3.3 & 13.8 \\
\textbf{P}    & 5.1 & 17.7 & 4.4 & 15.9 & 5.8 & 20.8 & 5.6 & 21.4 \\
\textbf{O}    & 10.2& 28.4 & 14.1& 37.8 & 11.3& 33.0 & 14.3& 38.3 \\
\textbf{OP}   & 12.9& 33.5 & 14.9& 39.8 & 14.6& 38.0 & 17.2& 44.1 \\
\textbf{OPT}  & 13.5& 35.6 & 16.4& 42.5 & 16.1& 40.2 & 18.8& 47.2 \\
\textbf{OPCT} & 13.3& 35.4 & 18.7& 45.9 & 16.1& 41.2 & 19.4& 48.9 \\
\textbf{ALL}  & 13.7& 34.7 & 19.0& 45.9 & 16.4& 41.0 & 19.6& 49.9 \\
\bottomrule
\end{tabular}%
}
\end{table}

\subsection{Detailed Inter-Subject Performance Analysis}
\label{app:backbone_analysis}

\Cref{tab:simon_results} presents a breakdown of performance across five different brain encoders. By examining both Top-1 and Top-5 metrics, we can discern nuanced differences in how each architecture handles spatial expansion and signal noise.

\textbf{Sensitivity to Frontal Noise Varies by Architecture.}
While the aggregate trend favors adding channels, the Top-5 data reveals architectural vulnerabilities. 
TSconv is remarkably robust; its Top-5 accuracy climbs steadily from 44.1\% (OP) to a peak of \textbf{49.9\%} (ALL). In contrast, \textbf{EEGProject} exhibits a "peaking" behavior similar to intra-subject models: it reaches its maximum efficacy at OPT (Top-5: 44.1\%) but suffers a decline when Central and Frontal channels are added (falling to 41.3\% at ALL). This suggests that EEGProject, likely optimized for specific frequency bands or spatial configurations, is less capable of filtering out the unrelated activity present in the anterior sensors compared to the multi-scale convolution approach of TSconv.

\textbf{Consistent Benefit for General-Purpose Backbones.}
For standard deep learning architectures like Shallownet and EEGnet, the inclusion of broader spatial data provides a clear stabilization effect. Shallownet sees its Top-5 accuracy improve from 35.9\% (OP) to 41.1\% (ALL). Even Deepnet, which struggles significantly in this sparse-data regime (Top-1 < 10\%), sees its Top-5 performance rise from 23.3\% to 25.6\% when using all channels. This confirms that the "broad spatial integration" hypothesis discussed in the main text is not specific to our SIMON method but is a fundamental characteristic of the inter-subject EEG retrieval task—more spatial context helps mitigate the domain shift, provided the encoder has sufficient capacity (like TSconv) to manage the increased dimensionality.

\begin{table}[h!]
\centering
\caption{Inter-subject classification results on SIMON dataset across different Brain Backbones and Channel combinations.}
\label{tab:simon_results}
\resizebox{0.8\textwidth}{!}{%
\begin{tabular}{l c c c c c c c c c c}
\toprule
\multirow{2}{*}{\textbf{Channel}} & \multicolumn{2}{c}{\textbf{EEGProject}} & \multicolumn{2}{c}{\textbf{TSconv}} & \multicolumn{2}{c}{\textbf{Shallownet}} & \multicolumn{2}{c}{\textbf{Deepnet}} & \multicolumn{2}{c}{\textbf{EEGnet}} \\
\cmidrule(lr){2-3} \cmidrule(lr){4-5} \cmidrule(lr){6-7} \cmidrule(lr){8-9} \cmidrule(lr){10-11}
 & Top-1 & Top-5 & Top-1 & Top-5 & Top-1 & Top-5 & Top-1 & Top-5 & Top-1 & Top-5 \\
\midrule
\textbf{F}    & 1.6 & 5.9  & 1.9 & 6.5  & 0.7 & 4.2  & 0.6 & 2.8 & 0.8 & 3.9  \\
\textbf{C}    & 2.5 & 9.8  & 2.7 & 10.0 & 2.2 & 9.4  & 1.2 & 5.2 & 2.4 & 10.3 \\
\textbf{T}    & 3.4 & 13.4 & 3.3 & 13.8 & 4.3 & 14.6 & 2.0 & 9.2 & 3.1 & 11.7 \\
\textbf{P}    & 6.0 & 20.9 & 5.6 & 21.4 & 5.1 & 20.1 & 3.5 & 14.1& 5.5 & 18.7 \\
\textbf{O}    & 13.3& 36.0 & 14.3& 38.3 & 10.8& 30.8 & 6.7 & 22.8& 10.9& 31.3 \\
\textbf{OP}   & 16.1& 41.0 & 17.2& 44.1 & 12.7& 35.9 & 7.0 & 23.3& 12.2& 36.0 \\
\textbf{OPT}  & 17.4& 44.1 & 18.8& 47.2 & 14.8& 39.1 & 8.0 & 25.5& 12.2& 38.5 \\
\textbf{OPCT} & 16.9& 43.5 & \textbf{19.4}& 48.9 & 14.8& 40.4 & 7.8 & 24.8& 12.7& 36.8 \\
\textbf{ALL}  & 16.4& 41.3 & \textbf{19.6}& \textbf{49.9} & 16.1 & 41.1 & 8.2 & 25.6& 12.0& 36.9 \\
\bottomrule
\end{tabular}%
}
\end{table}

\subsection{Scalability and Consistency Across Encoders}
\label{sec:generalization_heatmap}

\begin{figure}[H]
    \centering
    \begin{minipage}[t]{0.48\textwidth}
        \centering
        \includegraphics[width=\linewidth]{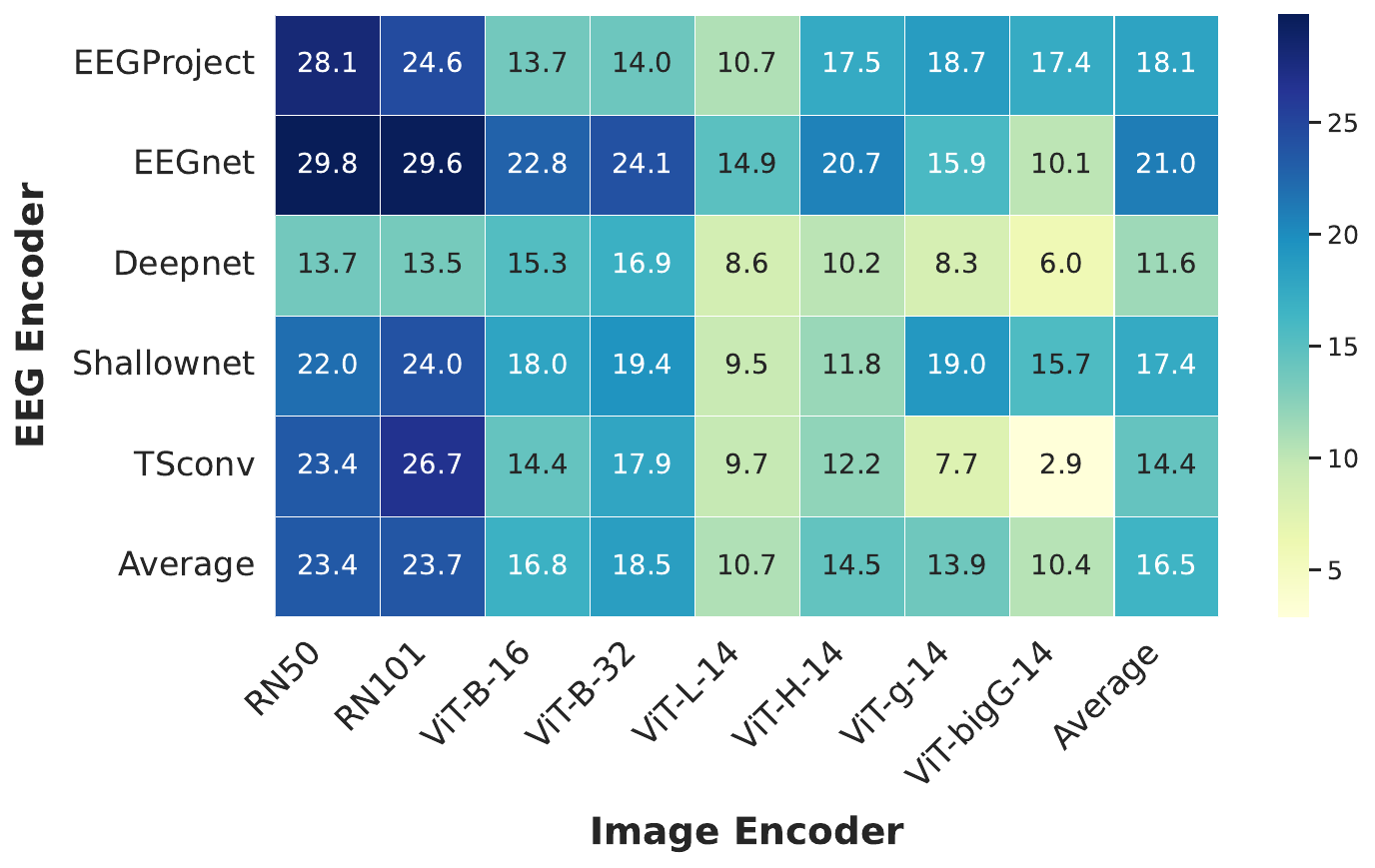}
        \caption{Top-1 accuracy improvement matrix (SIMON vs. Vanilla).}
        \label{fig:top1_improvement}
    \end{minipage}
    \hfill
    \begin{minipage}[t]{0.48\textwidth}
        \centering
        \includegraphics[width=\linewidth]{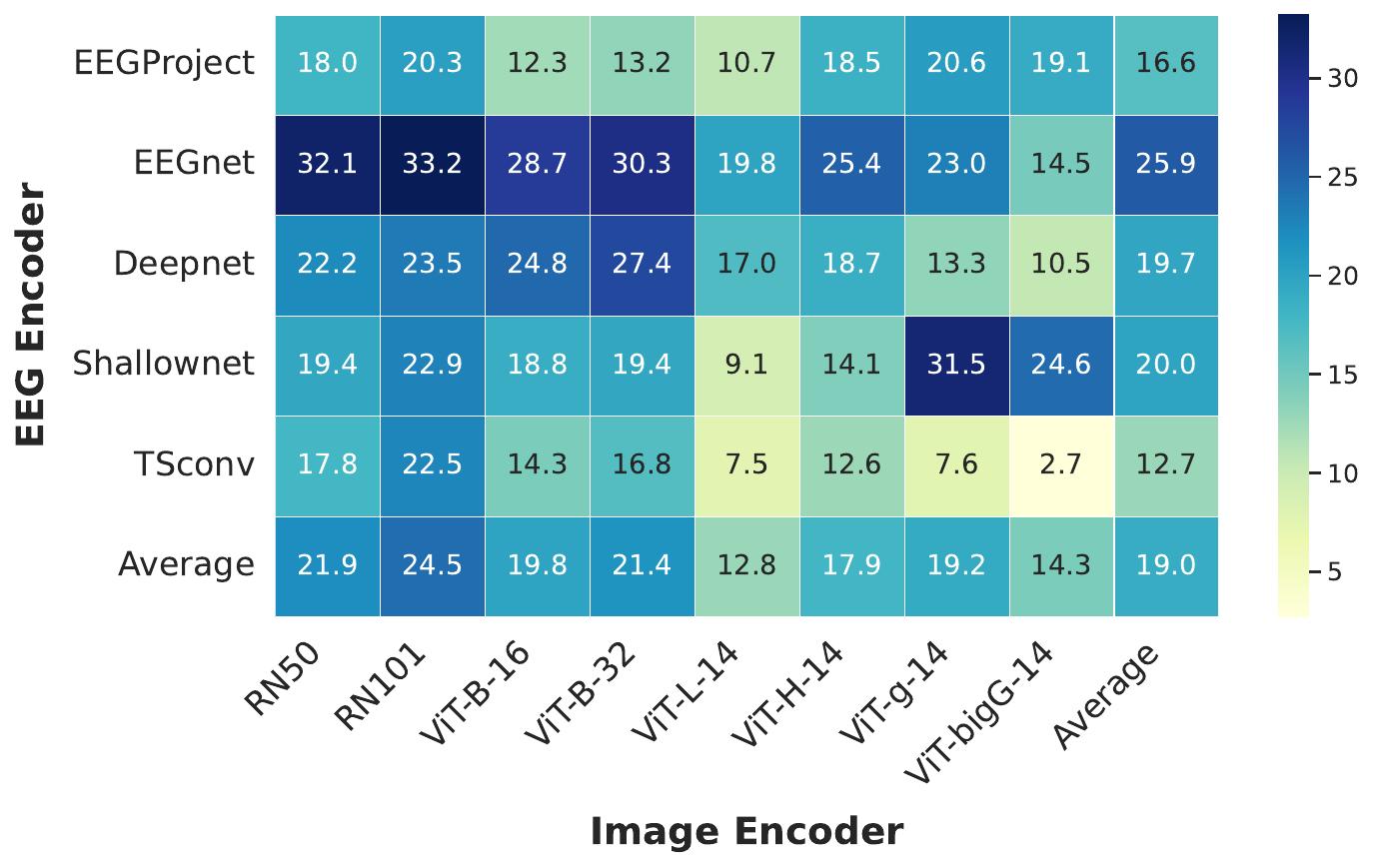}
        \caption{Top-5 accuracy improvement matrix (SIMON vs. Vanilla).}
        \label{fig:top5_improvement}
    \end{minipage}
\end{figure}

To assess the generalization capability of SIMON across diverse backbone configurations, we conducted a comprehensive cross-architecture benchmarking. We paired \textbf{five distinct brain encoders} (including EEGNet, TSconv, ATM, Gated-EEGProject, etc.) with \textbf{eight different vision backbones} (ranging from ResNet-50 to ViT-bigG-14). Figures~\ref{fig:top1_improvement} and \ref{fig:top5_improvement} visualize the performance gap between SIMON and the Vanilla baseline (standard contrastive learning without saliency guidance).

\textbf{Universal Improvement.} The results demonstrate an unequivocal improvement across all 40 architecture combinations. As detailed in our data, SIMON consistently outperforms the Vanilla baseline in both Top-1 and Top-5 accuracy. For standard CNNs like ResNet-50, we observe remarkable Top-1 gains reaching up to \textbf{+29.8\%} (from 21.7\% to 51.5\%) and Top-5 gains of \textbf{+32.1\%} (from 51.4\% to 83.5\%) under specific EEG encoder configurations. This indicates that saliency-aware integration adds a critical layer of semantic grounding that is absent in standard CNN feature maps.

\textbf{Scalability to Large Vision Transformers.} Crucially, this benefit extends to large-scale Vision Transformers, which typically struggle with data-scarce EEG alignment. Even with the massive ViT-g-14 backbone, SIMON achieves substantial improvements, boosting Top-1 accuracy by up to \textbf{+19.0\%} (from 7.4\% to 26.4\%) in challenging settings. Similarly, for ViT-bigG-14, we observe consistent gains across all brain encoders, with improvements ranging from +2.9\% to +17.4\%. This confirms that SIMON's ability to filter semantic distractors is robust and effective, regardless of the visual encoder's capacity or architecture type.

\subsection{Threshold Stability Analysis}
\label{app:tau_stability}

To examine the effect of the foreground threshold $\tau$ used in the saliency-aware sampling strategy, we conduct a dataset-wide stability analysis on the segmentation probability maps produced by BiRefNet over all 16,740 images in THINGS-EEG. Specifically, we measure how the foreground masks vary across a practical range of $\tau \in [0.3, 0.7]$.

We observe that the masks remain geometrically stable throughout this range. On average, only 1.1385\% of pixels fall within the uncertain probability interval. Moreover, the absolute mask area difference between $\tau=0.3$ and $\tau=0.7$ accounts for only 0.4463\% of the total image area, with a standard deviation of 0.9261\%.

These results indicate that varying $\tau$ changes only a negligible fraction of pixels. Consequently, the sampled foreground regions remain practically identical within the functional range used in our experiments, suggesting that the proposed sampling procedure is insensitive to moderate threshold variation.

\subsection{Necessity of Saliency Modeling}
\label{app:view_selection_comparison}

\begin{figure}[H]
    \centering
    \begin{subfigure}[b]{0.48\textwidth}
        \centering
        \includegraphics[width=0.8\linewidth]{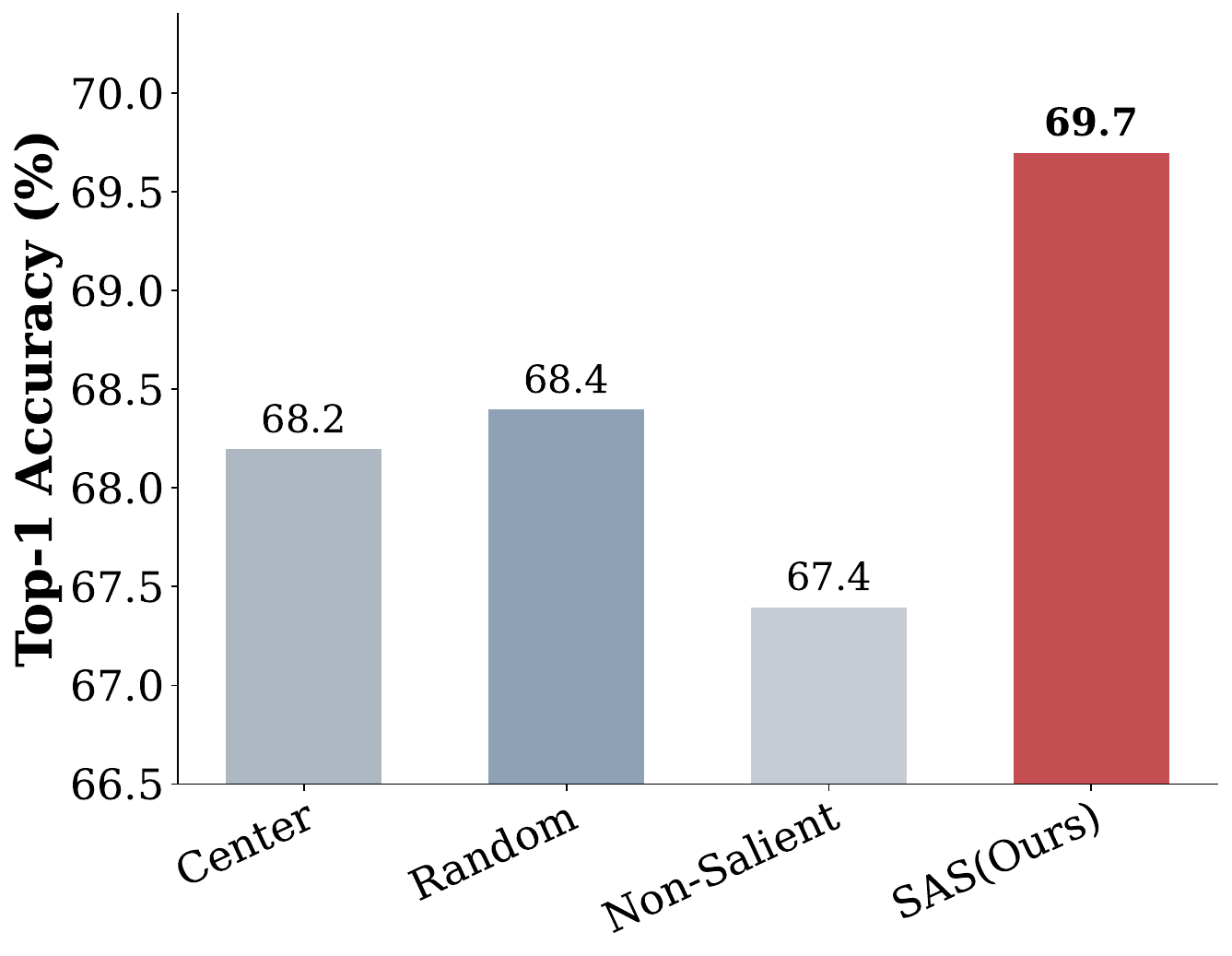}
        \caption{Top-1 accuracy}
        \label{fig:view_selection_comparison_top1_appendix}
    \end{subfigure}
    \hfill
    \begin{subfigure}[b]{0.48\textwidth}
        \centering
        \includegraphics[width=0.8\linewidth]{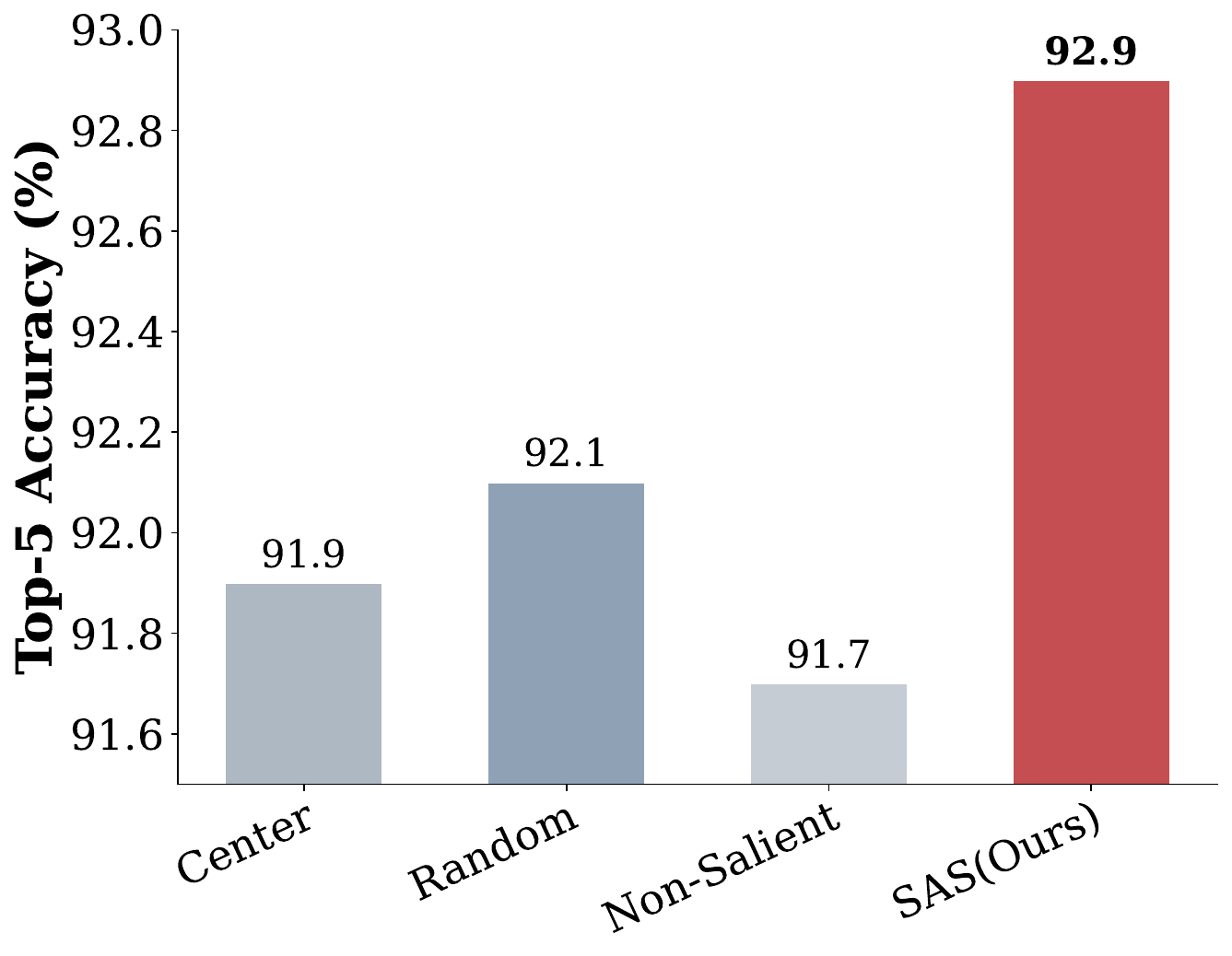}
        \caption{Top-5 accuracy}
        \label{fig:view_selection_comparison_top5_appendix}
    \end{subfigure}
    \caption{\textbf{Comparison of view-selection strategies under a fixed multi-view budget.}
    Geometric-center sampling serves as the center-fixed baseline. Random sampling produces only a minor change, whereas sampling from non-salient regions reduces retrieval accuracy. The SAS-based strategy yields the highest Top-1 and Top-5 performance under the same number of views.}
    \label{fig:view_selection_comparison_appendix}
\end{figure}

To test whether the gain comes specifically from saliency-aware selection, we compare four multi-view strategies under the same view budget: geometric-center sampling, random sampling, non-salient-region sampling, and the SAS-based strategy used in SIMON. Figure~\ref{fig:view_selection_comparison_appendix} reports the corresponding Top-1 and Top-5 results.

The comparison isolates the role of saliency modeling under matched multi-view settings. Random sampling remains close to the geometric-center baseline, whereas non-salient-region sampling reduces performance in both Top-1 and Top-5 accuracy. The strongest result is obtained by SIMON, whose sampled views are guided by foreground structure and visual saliency. This ordering indicates that the gain is not explained by multi-view augmentation alone, but depends on whether the selected views are directed toward semantically informative regions.

\subsection{Detailed ablation study on saliency-aware multi-view encoding}
\label{sec:detailed_three_factor_ablation}

Table~\ref{tab:three_factor_ablation_detailed} summarizes the retrieval performance across different configurations for both intra-subject and inter-subject tasks. These results are provided to evaluate the individual and joint contributions of saliency extraction, sampling strategies, and foveated view generation. The findings suggest that the performance gain is primarily driven by the synergy between these modules rather than the modification of any single factor.

Observations indicate that isolated adjustments—such as increasing view multiplicity without saliency guidance or applying foreground masking without saliency-aware sampling—yield inconsistent results or marginal fluctuations relative to the geometric-center baseline. The highest accuracy in both evaluation scenarios is consistently achieved when saliency-aware sampling is paired with multi-view generation under joint foreground and saliency guidance. This outcome points to the importance of integrating spatial priors directly into the foveated sampling and view generation process.

\begin{table}[htbp]
\centering
\caption{Detailed ablation results for saliency extraction, sampling strategy, and view generation on intra-subject and inter-subject tasks. ``N/A'' indicates settings in which saliency-aware sampling is undefined without saliency extraction.}
\label{tab:three_factor_ablation_detailed}
\begin{center}
\begin{small}
\setlength{\tabcolsep}{6pt} 
\resizebox{0.85\textwidth}{!}{
\begin{tabular}{lll cccc}
\toprule
\multirow{2}{*}{\textbf{Saliency Extraction}} & \multirow{2}{*}{\textbf{Sampling}} & \multirow{2}{*}{\textbf{View}} & \multicolumn{2}{c}{\textbf{Intra-subject (\%)}} & \multicolumn{2}{c}{\textbf{Inter-subject (\%)}} \\
\cmidrule(lr){4-5} \cmidrule(lr){6-7}
& & & \textbf{Top-1} & \textbf{Top-5} & \textbf{Top-1} & \textbf{Top-5} \\
\midrule
\multirow{4}{*}{None} & Random & Single & 68.2 & 91.9 & 18.9 & 48.9 \\
& Random & Multi & 68.4 & 92.1 & 19.4 & 49.5 \\
& Saliency-aware & Single & N/A & N/A & N/A & N/A \\
& Saliency-aware & Multi & N/A & N/A & N/A & N/A \\
\midrule
\multirow{4}{*}{Foreground} & Random & Single & 68.0 & 91.9 & 19.2 & 48.2 \\
& Random & Multi & 67.9 & 92.3 & 18.8 & 49.5 \\
& Saliency-aware & Single & 68.6 & 92.5 & 19.4 & 48.5 \\
& Saliency-aware & Multi & 69.1 & 92.2 & 19.4 & 49.5 \\
\midrule
\multirow{4}{*}{\begin{tabular}{@{}l@{}}Foreground \\ + Saliency\end{tabular}} & Random & Single & 67.9 & 92.2 & 19.2 & 49.7 \\
& Random & Multi & 68.5 & 92.0 & 19.1 & 49.5 \\
& Saliency-aware & Single & 68.2 & 92.1 & 19.5 & 48.6 \\
& Saliency-aware & Multi & \textbf{69.7} & \textbf{92.9} & \textbf{19.6} & \textbf{49.9} \\
\bottomrule
\end{tabular}
}
\end{small}
\end{center}
\end{table}

The consistency of this trend in the inter-subject task is a key observation. Despite the variability inherent in cross-subject EEG decoding and the lower absolute performance levels, the full configuration retains its position as the top-performing setting. This stability suggests that the foveated vision principles employed in the architecture help in extracting visual features that generalize across subjects. While the absolute margins reflect the low signal-to-noise ratio typical of non-invasive neural data, the alignment of the optimal configuration across both tasks indicates a robust architectural preference.

\subsection{Detailed Results for Visual Backbone Evaluation}
\label{appendix:backbones_full_results}

This section presents an extended evaluation of retrieval performance across different visual backbone configurations. Table~\ref{tab:backbone_ablation_top1_top5} reports the absolute Top-1 and Top-5 accuracy scores for all tested convolutional and Vision Transformer architectures, providing a direct assessment of the sensitivity of each method to the choice of visual feature extractor.

We compare the proposed SIMON framework with representative prior baselines under each experimental setting, using HyFI for intra-subject retrieval and NeuroBridge for inter-subject retrieval. The results show that SIMON maintains comparable or improved performance across different visual encoders, suggesting that its effectiveness is not tied to a particular backbone choice. The observed trends remain consistent under different visual feature parameterizations, providing additional support for the robustness of the proposed saliency-aware multi-view object-centric decoding framework.

\begin{table}[htbp]
\centering
\caption{Comparison of absolute Top-1 and Top-5 accuracy (\%) with different visual backbones.}
\label{tab:backbone_ablation_top1_top5}
\resizebox{0.85\linewidth}{!}{%
\begin{tabular}{lcccccccc}
\toprule
\multirow{3}{*}{\textbf{Backbone}} & \multicolumn{4}{c}{\textbf{Intra-subject}} & \multicolumn{4}{c}{\textbf{Inter-subject}} \\
\cmidrule(lr){2-5} \cmidrule(lr){6-9}
 & \multicolumn{2}{c}{\textbf{HyFI}} & \multicolumn{2}{c}{\textbf{SIMON}} & \multicolumn{2}{c}{\textbf{NeuroBridge}} & \multicolumn{2}{c}{\textbf{SIMON}} \\
\cmidrule(lr){2-3} \cmidrule(lr){4-5} \cmidrule(lr){6-7} \cmidrule(lr){8-9}
 & \textbf{Top-1} & \textbf{Top-5} & \textbf{Top-1} & \textbf{Top-5} & \textbf{Top-1} & \textbf{Top-5} & \textbf{Top-1} & \textbf{Top-5} \\
\midrule
RN50                           & 68.0 & 92.1 & 69.7 & 92.9 & 19.0 & 45.9 & 19.6 & 49.9 \\
RN101                          & 62.3 & 90.0 & 63.2 & 90.0 & 18.8 & 45.6 & 18.3 & 46.4 \\
ViT-H-14                       & 41.0 & 73.3 & 50.9 & 82.8 & 15.2 & 38.2 & 16.8 & 42.1 \\
ViT-L-14                       & 37.6 & 69.7 & 45.1 & 78.9 & 15.1 & 38.9 & 13.9 & 39.1 \\
ViT-bigG-14                    & 36.6 & 68.5 & 48.4 & 81.5 & 14.0 & 36.4 & 14.0 & 40.9 \\
ViT-g-14                       & 37.1 & 68.7 & 46.0 & 78.0 & 13.9 & 37.4 & 14.8 & 41.1 \\
\bottomrule
\end{tabular}%
}
\end{table}

\clearpage
\subsection{Qualitative Analysis}
\label{app:qualitative}

\begin{figure}[htbp!]
    \centering
    \setlength{\tabcolsep}{1pt} 
    \captionsetup{font=small,skip=2pt}
    \setlength{\tabcolsep}{0.5pt}
    \renewcommand{\arraystretch}{0.42}
    \setlength{\fboxrule}{0.8pt}
    \setlength{\fboxsep}{0pt}
    \setlength{\fboxrule}{1pt} 
    \setlength{\fboxsep}{0pt}  
    
    \begin{tabular}{cccccc}
        \small \textbf{Target} & \small \textbf{Top-1} & \small \textbf{Top-2} & \small \textbf{Top-3} & \small \textbf{Top-4} & \small \textbf{Top-5} \\
        \midrule
        
        \includegraphics[width=0.105\linewidth]{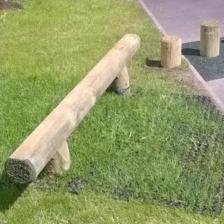} &
        \fcolorbox{red}{white}{\includegraphics[width=0.105\linewidth]{thingsimg/Good_Case/00004_balance_beam/Top1_balance_beam_04s.jpg}} &
        \includegraphics[width=0.105\linewidth]{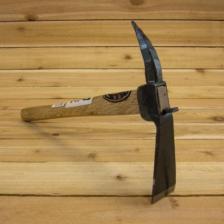} &
        \includegraphics[width=0.105\linewidth]{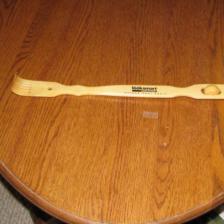} &
        \includegraphics[width=0.105\linewidth]{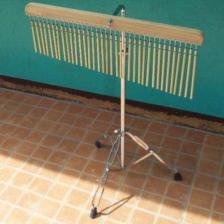} &
        \includegraphics[width=0.105\linewidth]{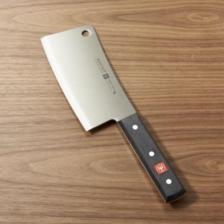} \\
        \addlinespace[1pt]
        
        \includegraphics[width=0.105\linewidth]{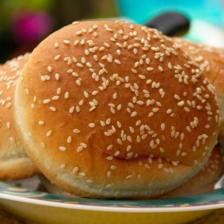} &
        \fcolorbox{red}{white}{\includegraphics[width=0.105\linewidth]{thingsimg/Good_Case/00027_bun/Top1_bun_01b.jpg}} &
        \includegraphics[width=0.105\linewidth]{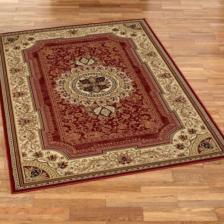} &
        \includegraphics[width=0.105\linewidth]{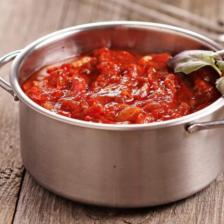} &
        \includegraphics[width=0.105\linewidth]{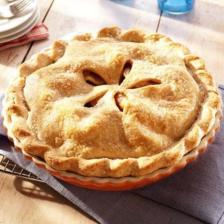} &
        \includegraphics[width=0.105\linewidth]{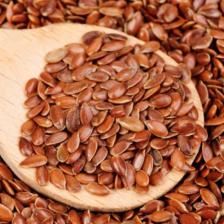} \\
        \addlinespace[1pt]

        \includegraphics[width=0.105\linewidth]{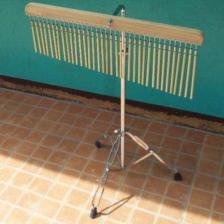} &
        \fcolorbox{red}{white}{\includegraphics[width=0.105\linewidth]{thingsimg/Good_Case/00041_chime/Top1_chime_08s.jpg}} &
        \includegraphics[width=0.105\linewidth]{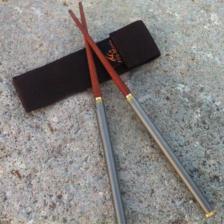} &
        \includegraphics[width=0.105\linewidth]{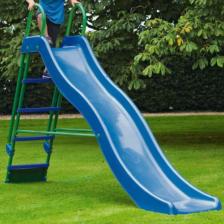} &
        \includegraphics[width=0.105\linewidth]{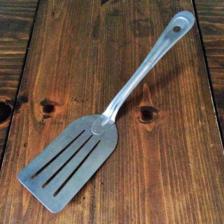} &
        \includegraphics[width=0.105\linewidth]{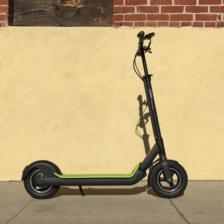} \\
        \addlinespace[1pt]

        \includegraphics[width=0.105\linewidth]{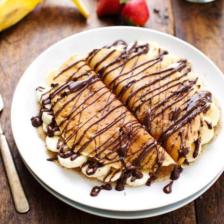} &
        \fcolorbox{red}{white}{\includegraphics[width=0.105\linewidth]{thingsimg/Good_Case/00055_crepe/Top1_crepe_18s.jpg}} &
        \includegraphics[width=0.105\linewidth]{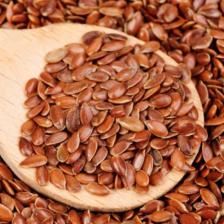} &
        \includegraphics[width=0.105\linewidth]{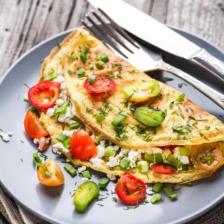} &
        \includegraphics[width=0.105\linewidth]{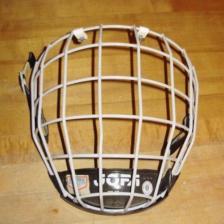} &
        \includegraphics[width=0.105\linewidth]{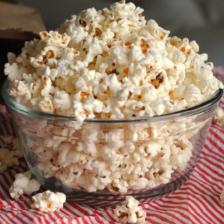} \\
        \addlinespace[1pt]

        \includegraphics[width=0.105\linewidth]{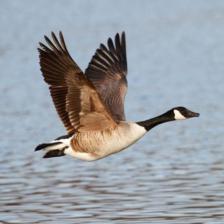} &
        \fcolorbox{red}{white}{\includegraphics[width=0.105\linewidth]{thingsimg/Good_Case/00086_goose/Top1_goose_02s.jpg}} &
        \includegraphics[width=0.105\linewidth]{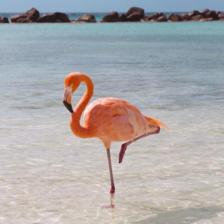} &
        \includegraphics[width=0.105\linewidth]{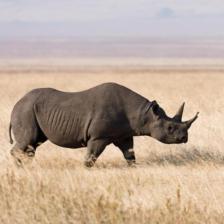} &
        \includegraphics[width=0.105\linewidth]{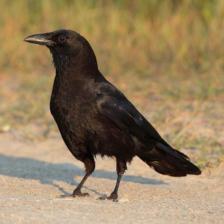} &
        \includegraphics[width=0.105\linewidth]{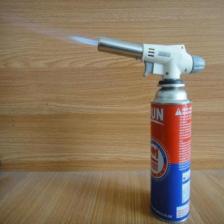} \\
        \addlinespace[1pt]

        \includegraphics[width=0.105\linewidth]{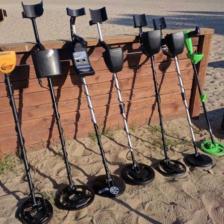} &
        \fcolorbox{red}{white}{\includegraphics[width=0.105\linewidth]{thingsimg/Good_Case/00114_metal_detector/Top1_metal_detector_02s.jpg}} &
        \includegraphics[width=0.105\linewidth]{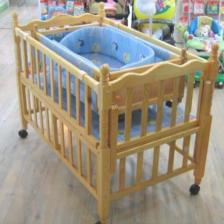} &
        \includegraphics[width=0.105\linewidth]{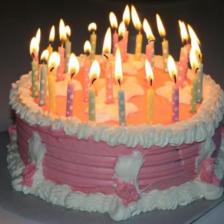} &
        \includegraphics[width=0.105\linewidth]{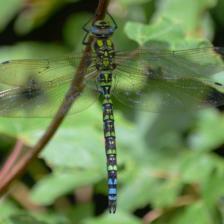} &
        \includegraphics[width=0.105\linewidth]{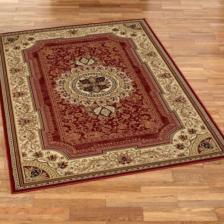} \\
        \addlinespace[1pt]

        \includegraphics[width=0.105\linewidth]{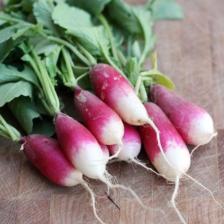} &
        \fcolorbox{red}{white}{\includegraphics[width=0.105\linewidth]{thingsimg/Good_Case/00147_radish/Top1_radish_10s.jpg}} &
        \includegraphics[width=0.105\linewidth]{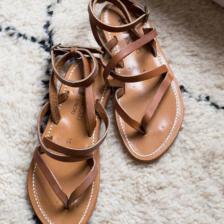} &
        \includegraphics[width=0.105\linewidth]{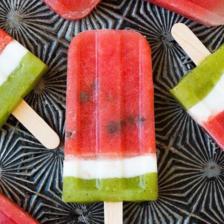} &
        \includegraphics[width=0.105\linewidth]{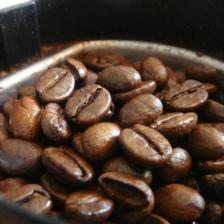} &
        \includegraphics[width=0.105\linewidth]{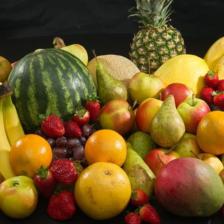} \\
        \addlinespace[1pt]

        \includegraphics[width=0.105\linewidth]{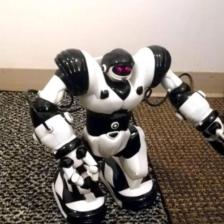} &
        \fcolorbox{red}{white}{\includegraphics[width=0.105\linewidth]{thingsimg/Good_Case/00151_robot/Top1_robot_02s.jpg}} &
        \includegraphics[width=0.105\linewidth]{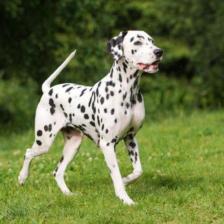} &
        \includegraphics[width=0.105\linewidth]{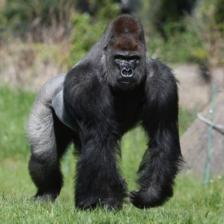} &
        \includegraphics[width=0.105\linewidth]{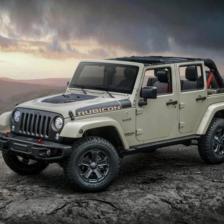} &
        \includegraphics[width=0.105\linewidth]{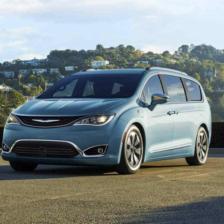} \\
        \addlinespace[1pt]

        \includegraphics[width=0.105\linewidth]{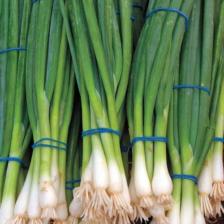} &
        \fcolorbox{red}{white}{\includegraphics[width=0.105\linewidth]{thingsimg/Good_Case/00158_scallion/Top1_scallion_04s.jpg}} &
        \includegraphics[width=0.105\linewidth]{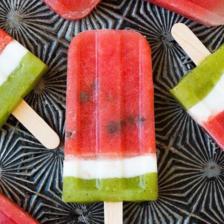} &
        \includegraphics[width=0.105\linewidth]{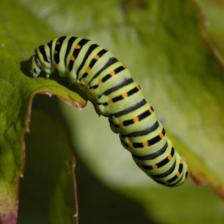} &
        \includegraphics[width=0.105\linewidth]{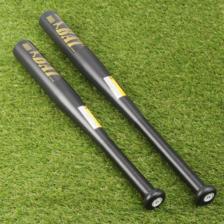} &
        \includegraphics[width=0.105\linewidth]{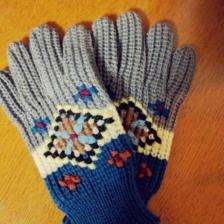} \\
        \addlinespace[1pt]

        \includegraphics[width=0.105\linewidth]{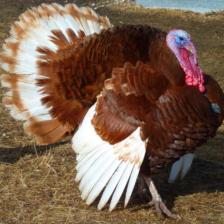} &
        \fcolorbox{red}{white}{\includegraphics[width=0.105\linewidth]{thingsimg/Good_Case/00190_turkey/Top1_turkey_01s.jpg}} &
        \includegraphics[width=0.105\linewidth]{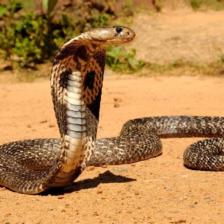} &
        \includegraphics[width=0.105\linewidth]{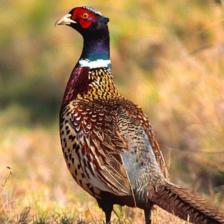} &
        \includegraphics[width=0.105\linewidth]{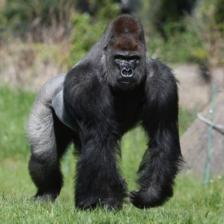} &
        \includegraphics[width=0.105\linewidth]{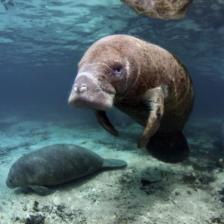} \\
        
    \end{tabular}
    \caption{\textbf{Qualitative visualization of successful retrieval cases.}}
    \label{fig:qualitative_results}
\end{figure}

\begin{figure}[p]
    \centering
    \setlength{\tabcolsep}{1pt} 
    \captionsetup{font=small,skip=2pt}
    \setlength{\tabcolsep}{0.5pt}
    \renewcommand{\arraystretch}{0.42}
    \setlength{\fboxrule}{0.8pt}
    \setlength{\fboxsep}{0pt}
    \setlength{\fboxrule}{1pt} 
    \setlength{\fboxsep}{0pt}
    
    \begin{tabular}{cccccc}
        \small \textbf{Target} & \small \textbf{Top-1} & \small \textbf{Top-2} & \small \textbf{Top-3} & \small \textbf{Top-4} & \small \textbf{Top-5} \\
        \midrule

        \includegraphics[width=0.105\linewidth]{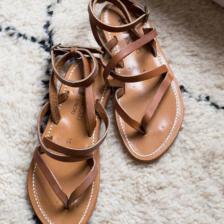} &
        \includegraphics[width=0.105\linewidth]{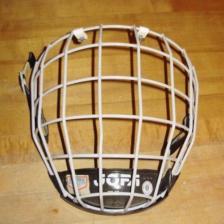} &
        \fcolorbox{red}{white}{\includegraphics[width=0.105\linewidth]{thingsimg/Bad_Case/00155_sandal/Top2_sandal_03s.jpg}} &
        \includegraphics[width=0.105\linewidth]{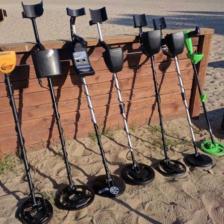} &
        \includegraphics[width=0.105\linewidth]{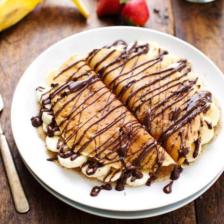} &
        \includegraphics[width=0.105\linewidth]{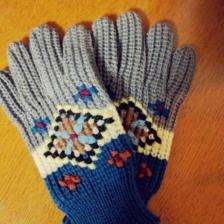} \\
        \addlinespace[1pt]

        \includegraphics[width=0.105\linewidth]{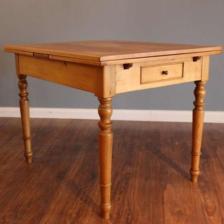} &
        \includegraphics[width=0.105\linewidth]{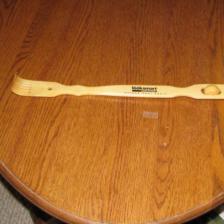} &
        \fcolorbox{red}{white}{\includegraphics[width=0.105\linewidth]{thingsimg/Bad_Case/00178_table/Top2_table_07s.jpg}} &
        \includegraphics[width=0.105\linewidth]{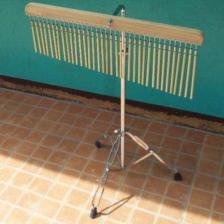} &
        \includegraphics[width=0.105\linewidth]{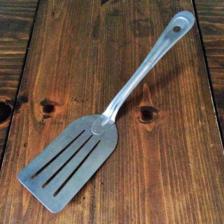} &
        \includegraphics[width=0.105\linewidth]{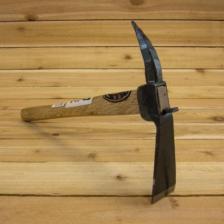} \\
        \addlinespace[1pt]
        
        \includegraphics[width=0.105\linewidth]{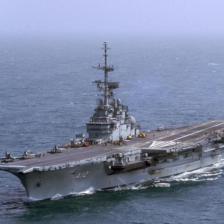} &
        \includegraphics[width=0.105\linewidth]{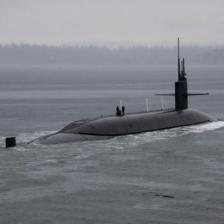} &
        \includegraphics[width=0.105\linewidth]{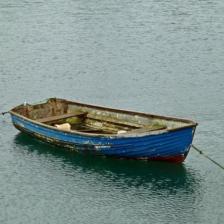} &
        \includegraphics[width=0.105\linewidth]{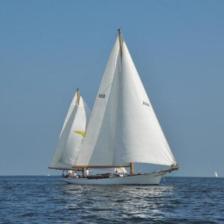} &
        \fcolorbox{red}{white}{\includegraphics[width=0.105\linewidth]{thingsimg/Bad_Case/00001_aircraft_carrier/Top4_aircraft_carrier_06s.jpg}} &
        \includegraphics[width=0.105\linewidth]{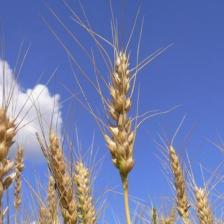} \\
        \addlinespace[1pt]

        \includegraphics[width=0.105\linewidth]{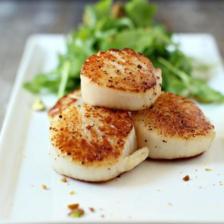} &
        \includegraphics[width=0.105\linewidth]{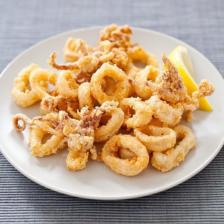} &
        \fcolorbox{red}{white}{\includegraphics[width=0.105\linewidth]{thingsimg/Bad_Case/00159_scallop/Top2_scallop_02s.jpg}} &
        \includegraphics[width=0.105\linewidth]{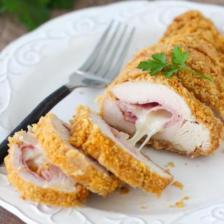} &
        \includegraphics[width=0.105\linewidth]{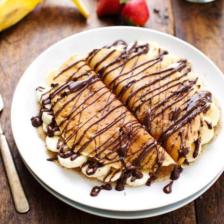} &
        \includegraphics[width=0.105\linewidth]{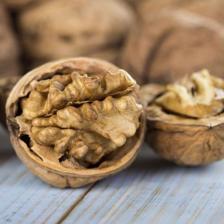} \\
        \addlinespace[1pt]
        
        \includegraphics[width=0.105\linewidth]{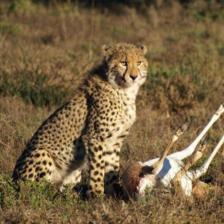} &
        \includegraphics[width=0.105\linewidth]{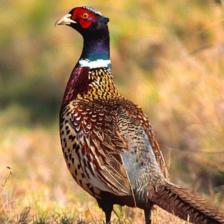} &
        \includegraphics[width=0.105\linewidth]{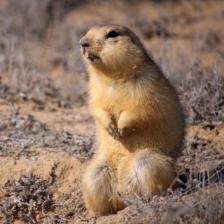} &
        \fcolorbox{red}{white}{\includegraphics[width=0.105\linewidth]{thingsimg/Bad_Case/00039_cheetah/Top3_cheetah_05s.jpg}} &
        \includegraphics[width=0.105\linewidth]{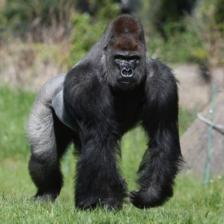} &
        \includegraphics[width=0.105\linewidth]{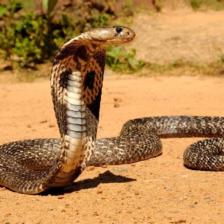} \\
        \addlinespace[1pt]

        \includegraphics[width=0.105\linewidth]{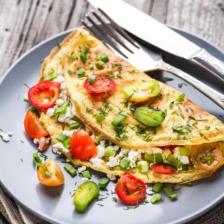} &
        \includegraphics[width=0.105\linewidth]{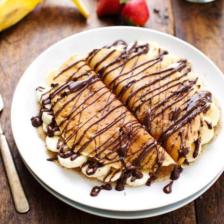} &
        \fcolorbox{red}{white}{\includegraphics[width=0.105\linewidth]{thingsimg/Bad_Case/00123_omelet/Top2_omelet_10s.jpg}} &
        \includegraphics[width=0.105\linewidth]{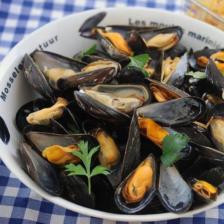} &
        \includegraphics[width=0.105\linewidth]{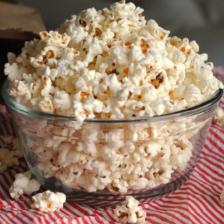} &
        \includegraphics[width=0.105\linewidth]{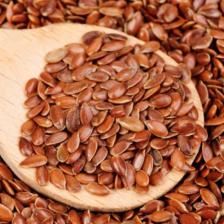} \\
        \addlinespace[1pt]

        \includegraphics[width=0.105\linewidth]{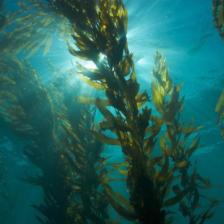} &
        \includegraphics[width=0.105\linewidth]{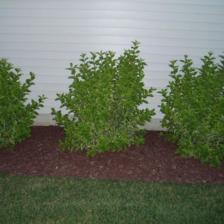} &
        \fcolorbox{red}{white}{\includegraphics[width=0.105\linewidth]{thingsimg/Bad_Case/00162_seaweed/Top2_seaweed_07s.jpg}} &
        \includegraphics[width=0.105\linewidth]{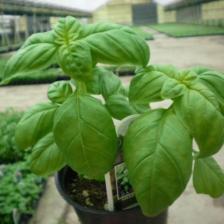} &
        \includegraphics[width=0.105\linewidth]{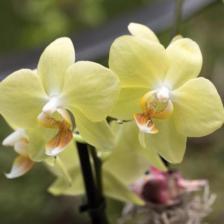} &
        \includegraphics[width=0.105\linewidth]{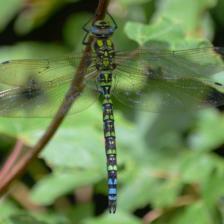} \\
        \addlinespace[1pt]

        \includegraphics[width=0.105\linewidth]{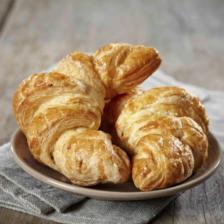} &
        \includegraphics[width=0.105\linewidth]{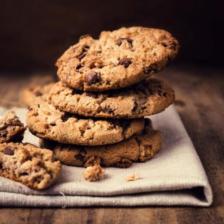} &
        \includegraphics[width=0.105\linewidth]{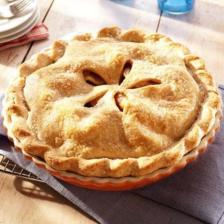} &
        \fcolorbox{red}{white}{\includegraphics[width=0.105\linewidth]{thingsimg/Bad_Case/00057_croissant/Top3_croissant_06s.jpg}} &
        \includegraphics[width=0.105\linewidth]{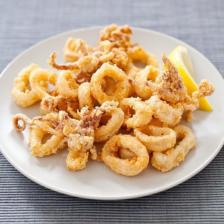} &
        \includegraphics[width=0.105\linewidth]{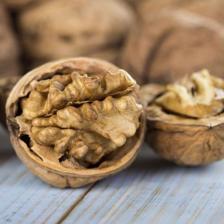} \\
        \addlinespace[1pt]

        \includegraphics[width=0.105\linewidth]{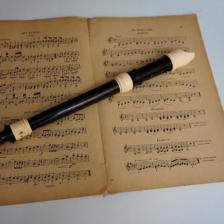} &
        \includegraphics[width=0.105\linewidth]{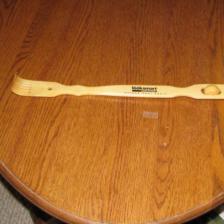} &
        \fcolorbox{red}{white}{\includegraphics[width=0.105\linewidth]{thingsimg/Bad_Case/00149_recorder/Top2_recorder_14s.jpg}} &
        \includegraphics[width=0.105\linewidth]{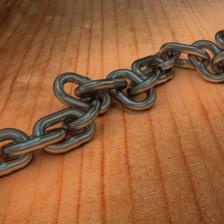} &
        \includegraphics[width=0.105\linewidth]{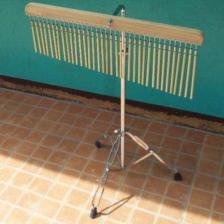} &
        \includegraphics[width=0.105\linewidth]{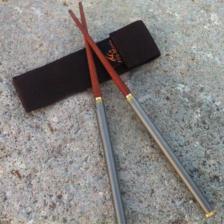} \\
        \addlinespace[1pt]

        \includegraphics[width=0.105\linewidth]{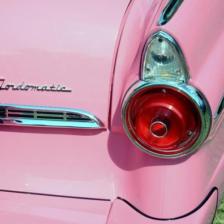} &
        \includegraphics[width=0.105\linewidth]{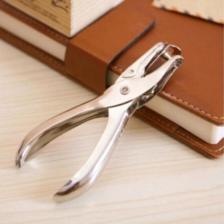} &
        \fcolorbox{red}{white}{\includegraphics[width=0.105\linewidth]{thingsimg/Bad_Case/00179_taillight/Top2_taillight_08s.jpg}} &
        \includegraphics[width=0.105\linewidth]{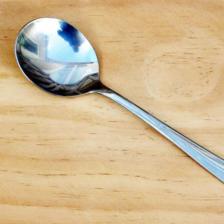} &
        \includegraphics[width=0.105\linewidth]{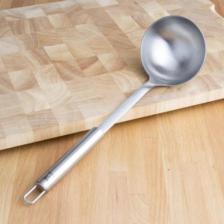} &
        \includegraphics[width=0.105\linewidth]{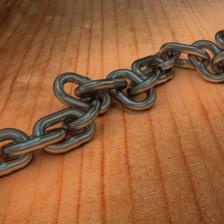} \\
        
    \end{tabular}
    \caption{\textbf{Qualitative visualization of failure cases.}}
    \label{fig:bad_cases}
\end{figure}

In this section, we present additional experimental results, showcasing both successful retrieval cases (Figure~\ref{fig:qualitative_results}) and failure cases (Figure~\ref{fig:bad_cases}). For each example, the Target column displays the query image (ground truth), while the subsequent columns show the Top-1 to Top-5 retrieved results. Furthermore, the correct match within the retrieved candidates is highlighted with a red bounding box to facilitate visualization.

\clearpage

\end{document}